\documentclass[10pt]{article} %
\usepackage[preprint]{tmlr}

\usepackage{amsmath,amsfonts,bm}

\def\matt#1{\textcolor{orange}{ #1}}
\def\harry#1{\textcolor{cyan}{ #1}}
\def\arg#1{\textcolor{blue}{ #1}}
\definecolor{gray1}{HTML}{7E7A77}

\newcommand{\figcenter}{{\em (Center)}}
\newcommand{\figright}{{\em (Right)}}
\newcommand{\figtop}{{\em (Top)}}
\newcommand{\figbottom}{{\em (Bottom)}}
\newcommand{\captiona}{{\em (a)}}
\newcommand{\captionb}{{\em (b)}}
\newcommand{\captionc}{{\em (c)}}
\newcommand{\captiond}{{\em (d)}}

\newcommand{\newterm}[1]{{\bf #1}}

\def\figref#1{figure~\ref{#1}}
\def\Figref#1{Figure~\ref{#1}}
\def\twofigref#1#2{figures \ref{#1} and \ref{#2}}
\def\quadfigref#1#2#3#4{figures \ref{#1}, \ref{#2}, \ref{#3} and \ref{#4}}
\def\secref#1{section~\ref{#1}}
\def\Secref#1{Section~\ref{#1}}
\def\twosecrefs#1#2{sections \ref{#1} and \ref{#2}}
\def\secrefs#1#2#3{sections \ref{#1}, \ref{#2} and \ref{#3}}
\def\eqref#1{equation~\ref{#1}}
\def\Eqref#1{Equation~\ref{#1}}
\def\plaineqref#1{\ref{#1}}
\def\chapref#1{chapter~\ref{#1}}
\def\Chapref#1{Chapter~\ref{#1}}
\def\rangechapref#1#2{chapters\ref{#1}--\ref{#2}}
\def\algref#1{algorithm~\ref{#1}}
\def\Algref#1{Algorithm~\ref{#1}}
\def\twoalgref#1#2{algorithms \ref{#1} and \ref{#2}}
\def\Twoalgref#1#2{Algorithms \ref{#1} and \ref{#2}}
\def\partref#1{part~\ref{#1}}
\def\Partref#1{Part~\ref{#1}}
\def\twopartref#1#2{parts \ref{#1} and \ref{#2}}

\def\ceil#1{\lceil #1 \rceil}
\def\floor#1{\lfloor #1 \rfloor}
\def\1{\bm{1}}
\newcommand{\train}{\mathcal{D}}
\newcommand{\valid}{\mathcal{D_{\mathrm{valid}}}}
\newcommand{\test}{\mathcal{D_{\mathrm{test}}}}

\def\eps{{\epsilon}}

\def\reta{{\textnormal{$\eta$}}}
\def\ra{{\textnormal{a}}}
\def\rb{{\textnormal{b}}}
\def\rc{{\textnormal{c}}}
\def\rd{{\textnormal{d}}}
\def\re{{\textnormal{e}}}
\def\rf{{\textnormal{f}}}
\def\rg{{\textnormal{g}}}
\def\rh{{\textnormal{h}}}
\def\ri{{\textnormal{i}}}
\def\rj{{\textnormal{j}}}
\def\rk{{\textnormal{k}}}
\def\rl{{\textnormal{l}}}
\def\rn{{\textnormal{n}}}
\def\ro{{\textnormal{o}}}
\def\rp{{\textnormal{p}}}
\def\rq{{\textnormal{q}}}
\def\rr{{\textnormal{r}}}
\def\rs{{\textnormal{s}}}
\def\rt{{\textnormal{t}}}
\def\ru{{\textnormal{u}}}
\def\rv{{\textnormal{v}}}
\def\rw{{\textnormal{w}}}
\def\rx{{\textnormal{x}}}
\def\ry{{\textnormal{y}}}
\def\rz{{\textnormal{z}}}

\def\rvepsilon{{\mathbf{\epsilon}}}
\def\rvtheta{{\mathbf{\theta}}}
\def\rva{{\mathbf{a}}}
\def\rvb{{\mathbf{b}}}
\def\rvc{{\mathbf{c}}}
\def\rvd{{\mathbf{d}}}
\def\rve{{\mathbf{e}}}
\def\rvf{{\mathbf{f}}}
\def\rvg{{\mathbf{g}}}
\def\rvh{{\mathbf{h}}}
\def\rvu{{\mathbf{i}}}
\def\rvj{{\mathbf{j}}}
\def\rvk{{\mathbf{k}}}
\def\rvl{{\mathbf{l}}}
\def\rvm{{\mathbf{m}}}
\def\rvn{{\mathbf{n}}}
\def\rvo{{\mathbf{o}}}
\def\rvp{{\mathbf{p}}}
\def\rvq{{\mathbf{q}}}
\def\rvr{{\mathbf{r}}}
\def\rvs{{\mathbf{s}}}
\def\rvt{{\mathbf{t}}}
\def\rvu{{\mathbf{u}}}
\def\rvv{{\mathbf{v}}}
\def\rvw{{\mathbf{w}}}
\def\rvx{{\mathbf{x}}}
\def\rvy{{\mathbf{y}}}
\def\rvz{{\mathbf{z}}}

\def\erva{{\textnormal{a}}}
\def\ervb{{\textnormal{b}}}
\def\ervc{{\textnormal{c}}}
\def\ervd{{\textnormal{d}}}
\def\erve{{\textnormal{e}}}
\def\ervf{{\textnormal{f}}}
\def\ervg{{\textnormal{g}}}
\def\ervh{{\textnormal{h}}}
\def\ervi{{\textnormal{i}}}
\def\ervj{{\textnormal{j}}}
\def\ervk{{\textnormal{k}}}
\def\ervl{{\textnormal{l}}}
\def\ervm{{\textnormal{m}}}
\def\ervn{{\textnormal{n}}}
\def\ervo{{\textnormal{o}}}
\def\ervp{{\textnormal{p}}}
\def\ervq{{\textnormal{q}}}
\def\ervr{{\textnormal{r}}}
\def\ervs{{\textnormal{s}}}
\def\ervt{{\textnormal{t}}}
\def\ervu{{\textnormal{u}}}
\def\ervv{{\textnormal{v}}}
\def\ervw{{\textnormal{w}}}
\def\ervx{{\textnormal{x}}}
\def\ervy{{\textnormal{y}}}
\def\ervz{{\textnormal{z}}}

\def\rmA{{\mathbf{A}}}
\def\rmB{{\mathbf{B}}}
\def\rmC{{\mathbf{C}}}
\def\rmD{{\mathbf{D}}}
\def\rmE{{\mathbf{E}}}
\def\rmF{{\mathbf{F}}}
\def\rmG{{\mathbf{G}}}
\def\rmH{{\mathbf{H}}}
\def\rmI{{\mathbf{I}}}
\def\rmJ{{\mathbf{J}}}
\def\rmK{{\mathbf{K}}}
\def\rmL{{\mathbf{L}}}
\def\rmM{{\mathbf{M}}}
\def\rmN{{\mathbf{N}}}
\def\rmO{{\mathbf{O}}}
\def\rmP{{\mathbf{P}}}
\def\rmQ{{\mathbf{Q}}}
\def\rmR{{\mathbf{R}}}
\def\rmS{{\mathbf{S}}}
\def\rmT{{\mathbf{T}}}
\def\rmU{{\mathbf{U}}}
\def\rmV{{\mathbf{V}}}
\def\rmW{{\mathbf{W}}}
\def\rmX{{\mathbf{X}}}
\def\rmY{{\mathbf{Y}}}
\def\rmZ{{\mathbf{Z}}}

\def\ermA{{\textnormal{A}}}
\def\ermB{{\textnormal{B}}}
\def\ermC{{\textnormal{C}}}
\def\ermD{{\textnormal{D}}}
\def\ermE{{\textnormal{E}}}
\def\ermF{{\textnormal{F}}}
\def\ermG{{\textnormal{G}}}
\def\ermH{{\textnormal{H}}}
\def\ermI{{\textnormal{I}}}
\def\ermJ{{\textnormal{J}}}
\def\ermK{{\textnormal{K}}}
\def\ermL{{\textnormal{L}}}
\def\ermM{{\textnormal{M}}}
\def\ermN{{\textnormal{N}}}
\def\ermO{{\textnormal{O}}}
\def\ermP{{\textnormal{P}}}
\def\ermQ{{\textnormal{Q}}}
\def\ermR{{\textnormal{R}}}
\def\ermS{{\textnormal{S}}}
\def\ermT{{\textnormal{T}}}
\def\ermU{{\textnormal{U}}}
\def\ermV{{\textnormal{V}}}
\def\ermW{{\textnormal{W}}}
\def\ermX{{\textnormal{X}}}
\def\ermY{{\textnormal{Y}}}
\def\ermZ{{\textnormal{Z}}}

\def\vzero{{\bm{0}}}
\def\vone{{\bm{1}}}
\def\vmu{{\bm{\mu}}}
\def\vtheta{{\bm{\theta}}}
\def\va{{\bm{a}}}
\def\vb{{\bm{b}}}
\def\vc{{\bm{c}}}
\def\vd{{\bm{d}}}
\def\ve{{\bm{e}}}
\def\vf{{\bm{f}}}
\def\vg{{\bm{g}}}
\def\vh{{\bm{h}}}
\def\vi{{\bm{i}}}
\def\vj{{\bm{j}}}
\def\vk{{\bm{k}}}
\def\vl{{\bm{l}}}
\def\vm{{\bm{m}}}
\def\vn{{\bm{n}}}
\def\vo{{\bm{o}}}
\def\vp{{\bm{p}}}
\def\vq{{\bm{q}}}
\def\vr{{\bm{r}}}
\def\vs{{\bm{s}}}
\def\vt{{\bm{t}}}
\def\vu{{\bm{u}}}
\def\vv{{\bm{v}}}
\def\vw{{\bm{w}}}
\def\vx{{\bm{x}}}
\def\vy{{\bm{y}}}
\def\vz{{\bm{z}}}

\def\evalpha{{\alpha}}
\def\evbeta{{\beta}}
\def\evepsilon{{\epsilon}}
\def\evlambda{{\lambda}}
\def\evomega{{\omega}}
\def\evmu{{\mu}}
\def\evpsi{{\psi}}
\def\evsigma{{\sigma}}
\def\evtheta{{\theta}}
\def\eva{{a}}
\def\evb{{b}}
\def\evc{{c}}
\def\evd{{d}}
\def\eve{{e}}
\def\evf{{f}}
\def\evg{{g}}
\def\evh{{h}}
\def\evi{{i}}
\def\evj{{j}}
\def\evk{{k}}
\def\evl{{l}}
\def\evm{{m}}
\def\evn{{n}}
\def\evo{{o}}
\def\evp{{p}}
\def\evq{{q}}
\def\evr{{r}}
\def\evs{{s}}
\def\evt{{t}}
\def\evu{{u}}
\def\evv{{v}}
\def\evw{{w}}
\def\evx{{x}}
\def\evy{{y}}
\def\evz{{z}}

\def\mA{{\bm{A}}}
\def\mB{{\bm{B}}}
\def\mC{{\bm{C}}}
\def\mD{{\bm{D}}}
\def\mE{{\bm{E}}}
\def\mF{{\bm{F}}}
\def\mG{{\bm{G}}}
\def\mH{{\bm{H}}}
\def\mI{{\bm{I}}}
\def\mJ{{\bm{J}}}
\def\mK{{\bm{K}}}
\def\mL{{\bm{L}}}
\def\mM{{\bm{M}}}
\def\mN{{\bm{N}}}
\def\mO{{\bm{O}}}
\def\mP{{\bm{P}}}
\def\mQ{{\bm{Q}}}
\def\mR{{\bm{R}}}
\def\mS{{\bm{S}}}
\def\mT{{\bm{T}}}
\def\mU{{\bm{U}}}
\def\mV{{\bm{V}}}
\def\mW{{\bm{W}}}
\def\mX{{\bm{X}}}
\def\mY{{\bm{Y}}}
\def\mZ{{\bm{Z}}}
\def\mBeta{{\bm{\beta}}}
\def\mPhi{{\bm{\Phi}}}
\def\mLambda{{\bm{\Lambda}}}
\def\mSigma{{\bm{\Sigma}}}

\DeclareMathAlphabet{\mathsfit}{\encodingdefault}{\sfdefault}{m}{sl}
\SetMathAlphabet{\mathsfit}{bold}{\encodingdefault}{\sfdefault}{bx}{n}
\newcommand{\tens}[1]{\bm{\mathsfit{#1}}}
\def\tA{{\tens{A}}}
\def\tB{{\tens{B}}}
\def\tC{{\tens{C}}}
\def\tD{{\tens{D}}}
\def\tE{{\tens{E}}}
\def\tF{{\tens{F}}}
\def\tG{{\tens{G}}}
\def\tH{{\tens{H}}}
\def\tI{{\tens{I}}}
\def\tJ{{\tens{J}}}
\def\tK{{\tens{K}}}
\def\tL{{\tens{L}}}
\def\tM{{\tens{M}}}
\def\tN{{\tens{N}}}
\def\tO{{\tens{O}}}
\def\tP{{\tens{P}}}
\def\tQ{{\tens{Q}}}
\def\tR{{\tens{R}}}
\def\tS{{\tens{S}}}
\def\tT{{\tens{T}}}
\def\tU{{\tens{U}}}
\def\tV{{\tens{V}}}
\def\tW{{\tens{W}}}
\def\tX{{\tens{X}}}
\def\tY{{\tens{Y}}}
\def\tZ{{\tens{Z}}}

\def\gA{{\mathcal{A}}}
\def\gB{{\mathcal{B}}}
\def\gC{{\mathcal{C}}}
\def\gD{{\mathcal{D}}}
\def\gE{{\mathcal{E}}}
\def\gF{{\mathcal{F}}}
\def\gG{{\mathcal{G}}}
\def\gH{{\mathcal{H}}}
\def\gI{{\mathcal{I}}}
\def\gJ{{\mathcal{J}}}
\def\gK{{\mathcal{K}}}
\def\gL{{\mathcal{L}}}
\def\gM{{\mathcal{M}}}
\def\gN{{\mathcal{N}}}
\def\gO{{\mathcal{O}}}
\def\gP{{\mathcal{P}}}
\def\gQ{{\mathcal{Q}}}
\def\gR{{\mathcal{R}}}
\def\gS{{\mathcal{S}}}
\def\gT{{\mathcal{T}}}
\def\gU{{\mathcal{U}}}
\def\gV{{\mathcal{V}}}
\def\gW{{\mathcal{W}}}
\def\gX{{\mathcal{X}}}
\def\gY{{\mathcal{Y}}}
\def\gZ{{\mathcal{Z}}}

\def\sA{{\mathbb{A}}}
\def\sB{{\mathbb{B}}}
\def\sC{{\mathbb{C}}}
\def\sD{{\mathbb{D}}}
\def\sF{{\mathbb{F}}}
\def\sG{{\mathbb{G}}}
\def\sH{{\mathbb{H}}}
\def\sI{{\mathbb{I}}}
\def\sJ{{\mathbb{J}}}
\def\sK{{\mathbb{K}}}
\def\sL{{\mathbb{L}}}
\def\sM{{\mathbb{M}}}
\def\sN{{\mathbb{N}}}
\def\sO{{\mathbb{O}}}
\def\sP{{\mathbb{P}}}
\def\sQ{{\mathbb{Q}}}
\def\sR{{\mathbb{R}}}
\def\sS{{\mathbb{S}}}
\def\sT{{\mathbb{T}}}
\def\sU{{\mathbb{U}}}
\def\sV{{\mathbb{V}}}
\def\sW{{\mathbb{W}}}
\def\sX{{\mathbb{X}}}
\def\sY{{\mathbb{Y}}}
\def\sZ{{\mathbb{Z}}}

\def\emLambda{{\Lambda}}
\def\emA{{A}}
\def\emB{{B}}
\def\emC{{C}}
\def\emD{{D}}
\def\emE{{E}}
\def\emF{{F}}
\def\emG{{G}}
\def\emH{{H}}
\def\emI{{I}}
\def\emJ{{J}}
\def\emK{{K}}
\def\emL{{L}}
\def\emM{{M}}
\def\emN{{N}}
\def\emO{{O}}
\def\emP{{P}}
\def\emQ{{Q}}
\def\emR{{R}}
\def\emS{{S}}
\def\emT{{T}}
\def\emU{{U}}
\def\emV{{V}}
\def\emW{{W}}
\def\emX{{X}}
\def\emY{{Y}}
\def\emZ{{Z}}
\def\emSigma{{\Sigma}}

\newcommand{\etens}[1]{\mathsfit{#1}}
\def\etLambda{{\etens{\Lambda}}}
\def\etA{{\etens{A}}}
\def\etB{{\etens{B}}}
\def\etC{{\etens{C}}}
\def\etD{{\etens{D}}}
\def\etE{{\etens{E}}}
\def\etF{{\etens{F}}}
\def\etG{{\etens{G}}}
\def\etH{{\etens{H}}}
\def\etI{{\etens{I}}}
\def\etJ{{\etens{J}}}
\def\etK{{\etens{K}}}
\def\etL{{\etens{L}}}
\def\etM{{\etens{M}}}
\def\etN{{\etens{N}}}
\def\etO{{\etens{O}}}
\def\etP{{\etens{P}}}
\def\etQ{{\etens{Q}}}
\def\etR{{\etens{R}}}
\def\etS{{\etens{S}}}
\def\etT{{\etens{T}}}
\def\etU{{\etens{U}}}
\def\etV{{\etens{V}}}
\def\etW{{\etens{W}}}
\def\etX{{\etens{X}}}
\def\etY{{\etens{Y}}}
\def\etZ{{\etens{Z}}}

\newcommand{\pdata}{p_{\rm{data}}}
\newcommand{\ptrain}{\hat{p}_{\rm{data}}}
\newcommand{\Ptrain}{\hat{P}_{\rm{data}}}
\newcommand{\pmodel}{p_{\rm{model}}}
\newcommand{\Pmodel}{P_{\rm{model}}}
\newcommand{\ptildemodel}{\tilde{p}_{\rm{model}}}
\newcommand{\pencode}{p_{\rm{encoder}}}
\newcommand{\pdecode}{p_{\rm{decoder}}}
\newcommand{\precons}{p_{\rm{reconstruct}}}

\newcommand{\laplace}{\mathrm{Laplace}} %

\newcommand{\E}{\mathbb{E}}
\newcommand{\Ls}{\mathcal{L}}
\newcommand{\R}{\mathbb{R}}
\newcommand{\emp}{\tilde{p}}
\newcommand{\lr}{\alpha}
\newcommand{\reg}{\lambda}
\newcommand{\rect}{\mathrm{rectifier}}
\newcommand{\softmax}{\mathrm{softmax}}
\newcommand{\sigmoid}{\sigma}
\newcommand{\softplus}{\zeta}
\newcommand{\KL}{D_{\mathrm{KL}}}
\newcommand{\Var}{\mathrm{Var}}
\newcommand{\standarderror}{\mathrm{SE}}
\newcommand{\Cov}{\mathrm{Cov}}
\newcommand{\normlzero}{L^0}
\newcommand{\normlone}{L^1}
\newcommand{\normltwo}{L^2}
\newcommand{\normlp}{L^p}
\newcommand{\normmax}{L^\infty}
\newcommand{\norm}[1]{\left\lVert#1\right\rVert}
\newcommand{\parents}{Pa} %

\DeclareMathOperator*{\argmax}{arg\,max}
\DeclareMathOperator*{\argmin}{arg\,min}

\DeclareMathOperator{\sign}{sign}
\DeclareMathOperator{\Tr}{Tr}
\let\ab\allowbreak

\usepackage{amsthm}
\usepackage{thmtools, thm-restate}
\usepackage{xcolor}
\usepackage[normalem]{ulem} %
\usepackage{float}
\usepackage{xr}
\usepackage{balance}
\usepackage{xspace}
\usepackage{url}
\usepackage{expl3}
\usepackage{dsfont}
\usepackage{bbm}
\usepackage{grffile}
\usepackage{tikz}
\usepackage{float}
\usepackage{svg}
\usepackage{subcaption}
\usepackage{relsize}  %
\usepackage{makecell}
\usepackage{refcount}
\usepackage{xstring}
\usepackage{multirow}
\usepackage{natbib}
\usepackage{rotating}
\usepackage{comment}
\usepackage{adjustbox}
\usepackage{calc}
\usepackage{stmaryrd}
\usepackage{standalone}
\usepackage{balance}
\usepackage{makecell}
\usepackage[utf8]{inputenc} %
\usepackage[T1]{fontenc}    %
\usepackage{hyperref}       %
\usepackage[capitalise]{cleveref}
\usepackage{url}            %
\usepackage{amsfonts}       %
\usepackage{amssymb}
\usepackage{amsmath}    %
\usepackage{mathtools}  %
\usepackage{nicefrac}       %
\usepackage{microtype}      %
\usepackage{xcolor}         %
\usepackage[table]{xcolor}
\usepackage{lmodern}
\usepackage{enumitem}       %
\usepackage{bold-extra}
\usepackage{microtype}
\usepackage{graphicx}
\usepackage{caption}

\usepackage[linesnumbered,ruled,algo2e]{algorithm2e}
\newcommand{\theHalgorithm}{\arabic{algorithm}}
\usepackage{algpseudocode}
\usepackage{algorithm}
\usepackage{algcompatible}

\renewcommand{\arraystretch}{1.1} %
\newcommand{\bestcell}[1]{\cellcolor{gray!30}\textbf{#1}}
\newcommand{\Q}{\mathbf{Q}}
\newcommand{\q}{q}

\definecolor{blue1}{HTML}{2851CC}
\definecolor{red1}{HTML}{E00000}

\hypersetup{
    colorlinks,
    linkcolor = red1,
    citecolor = blue1,
    urlcolor={blue!80!black},
}

\newtheorem{definition}{Definition}[section]
\newtheorem{remark}{Remark}[section]

\newtheorem{corollary}{Corollary}[section]

\def\matt#1{\textcolor{red}{ #1}}
\def\harry#1{\textcolor{blue}{ #1}}
\def\arg#1{\textcolor{magenta}{ #1}}

\newcommand{\GSC}{GSC\xspace}
\newcommand{\GSCn}{GSC$_{\text{n}}$\xspace}
\newcommand{\GSCun}{GSC$_{\text{un}}$\xspace}
\newcommand{\GDE}{GDE\xspace}
\newcommand{\GDEf}{$\text{GDE}(f)$}
\newcommand{\D}[2]{\mathcal{D}^2_{#1}(#2)}
\newcommand{\Dnorm}[2]{\overline{\mathcal{D}}_{#1}^2(#2)}

\newcommand{\digraph}{digraph\xspace}
\newcommand{\digraphs}{digraphs\xspace}

\renewcommand{\dots}{...}
\renewcommand{\ldots}{...}

\newcommand{\diag}[1]{\operatorname{diag}(#1)}

\renewcommand*{\top}{{\mkern-1.4mu\mathsf{T}}}
\newcommand{\chivec}[1] {{\chi_{\raisebox{-0.15em}{\scalebox{0.6}{$#1$}}}}}

\renewcommand{\eqref}[1]{\ref{#1}}
\newcommand{\TODO}[1] 			{{\color{red} #1}}
\newcommand{\NEW}[1] 				{{\color{blue} #1}}
\usepackage[normalem]{ulem}
\newcommand{\RMV}[1] 	  {{\color{orange} \sout{#1}}}
\newcommand{\NOTE}[1] 				{{\color{red} #1}}

\newcommand{\algo}[1] {\text{#1}}

\newcommand{\inlinetitle}[2]{\noindent\textbf{\textit{#1}{#2}}~~}

\newcommand{\ie}   			{i.e.\@\xspace}
\newcommand{\eg}   			{e.g.\@\xspace}
\newcommand{\padded}[1] {\,#1\,}
\newcommand{\mydef}			{:=}
\newcommand{\wrt}   		{w.r.t.\@\xspace}
\newcommand{\st}   			{s.t.\@\xspace}

\newcommand{\one}   		{\boldsymbol{1}}

\newcommand\simtilde[1]{\mathbin{%
    \stackrel{\sim}{\smash{#1}\rule{0pt}{1.05ex}}%
    }}
\newcommand\simtildesmall[1]{\mathbin{%
    \stackrel{\sim}{\smash{#1}\rule{0pt}{0.65ex}}%
    }}

\newcommand{\nuplusxi}

\newcommand{\bestpar}[2] {(#1)^*_{\textsc{#2}}}
\newcommand{\hyperp}[1]{\tiny{\tiny(#1)}}

\title{Generalized Dirichlet Energy and Graph Laplacians \\for Clustering Directed and Undirected Graphs}

\author{\vspace{1em}\\
 \name Harry Sevi\,$^\star$ \email harry.sevi@protonmail.com
      \vspace{-.8em}
			\AND
			\name Gwendal Debaussart-Joniec\,$^\star$ \email gwendal.debaussart@ens-paris-saclay.fr
      \vspace{-.8em}
			\AND
      \name Malik Hacini\,$^{\star\diamond}$
			\email malik.hacini@ens-paris-saclay.fr
			\vspace{-.8em}
			\AND
      \name Matthieu Jonckheere\,$^\circ$ \email matthieu.jonckheere@laas.fr
			\vspace{-.8em}
			\AND
      \name Argyris Kalogeratos\,$^\star$  \email argyris.kalogeratos@ens-paris-saclay.fr\ \vspace{1em}\\
      \addr $^\star$\,ENS Paris-Saclay, Universit\'e Paris-Saclay, 			CNRS, Centre Borelli, F-91190 Gif-sur-Yvette, France. \\
			$^\diamond$\,ENSIMAG, Grenoble INP, Université Grenoble Alpes, F-38400 Saint Martin d'Hères, France.\\
			$^\circ$\,CNRS, Laboratory for Analysis and Architecture of Systems, F-31400 Toulouse, France.
}

\def\month{MM}  %
\def\year{YYYY} %
\def\openreview{\url{https://openreview.net/forum?id=XXXX}} %

\begin{document}

\maketitle
\begin{abstract}
Clustering in directed graphs remains a fundamental challenge due to the asymmetry in edge connectivity, which limits the applicability of classical spectral methods originally designed for undirected graphs. A common workaround is to symmetrize the adjacency matrix, but this often leads to losing critical directional information.
In this work, we introduce the \emph{generalized Dirichlet energy} (GDE), a novel energy functional that extends the classical Dirichlet energy to handle arbitrary positive vertex measures and Markov transition matrices. GDE provides a unified framework applicable to both directed and undirected graphs, and is closely tied to the diffusion dynamics of random walks.
Building on this framework, we propose the \emph{generalized spectral clustering} (GSC) method that enables the principled clustering of weakly connected digraphs without resorting to the introduction of teleportation to the random walk transition matrix. A key component of our approach is the utilization of a parametrized vertex measure encoding graph directionality and density. %
Experiments on real-world point-cloud datasets demonstrate that GSC consistently outperforms existing spectral clustering approaches in terms of clustering accuracy and robustness, offering a powerful new tool for graph-based data analysis.

\end{abstract}

\section{Introduction} \label{sect:intro}
Clustering is a fundamental tool broadly used to uncover structure in large and complex datasets. In many modern applications, such as web graphs, citation networks, and information diffusion in social or transportation systems, a dataset is naturally represented as a directed graph (digraph), where edge direction encodes asymmetric relationships between data objects. Among the several graph clustering methods, spectral clustering has become popular due to its conceptual simplicity, efficiency, and strong theoretical underpinnings \citep{ng2002spectral, peng2015partitioning, boedihardjo2021performance}. A foundational concept in spectral clustering is the \emph{Dirichlet energy}, which quantifies the smoothness of a function over its domain. For a graph function, defined over the vertices of a graph, this measures the variability of the function across adjacent vertices, hence capturing local regularity with respect to graph connectivity. This notion underlies many successful clustering algorithms that have appeared in machine learning, network science, and other domains.

With data represented as a graph, graph clustering aims at partitioning that graph into subsets with high internal connectivity and sparse external links. This task is typically formulated as a discrete optimization problem, such as minimizing the graph cut or normalized cut, whose relaxed form corresponds to minimizing the Dirichlet energy of a graph function associated with a Laplacian operator \citep{von2007tutorial, shi2000normalized}. Spectral clustering solves this relaxed problem by computing the leading eigenvectors of the Laplacian, yielding low-dimensional embeddings that can be clustered via standard methods such as $k$-means.

Most spectral clustering methods are designed for undirected graphs. Many real-world systems, though, such as social and content networks or flow-based systems like transportation, are best modeled as digraphs carrying critical information in edge directionality.
A widely adopted practice in handling weakly connected digraphs in graph machine learning tasks, such as spectral clustering \citep{zhou2005learning} or node classification \citep{peach2019semi}, is the routine reliance on the teleporting random walk \citep{page1999pagerank}. This process guarantees ergodicity, hence it is often used as a substitute for the natural random walk that may not be ergodic in directed graphs. Despite offering an alignment with classical ergodic-based theoretical frameworks, the teleporting random walk constitutes a convenient workaround rather than a principled modeling decision. Some of the drawbacks it introduces are: First, it adds uniform transitions to all vertices, as it blends the original graph structure with that of a complete graph. This alters its topology and injects non-local interactions, potentially masking critical structural information of the original graph. Imposing ergodicity in cases where it does not naturally hold can lead to misleading conclusions about the true graph dynamics \citep{schaub2019multiscale}. Moreover, by densifying the graph it increases substantially the computational cost in practice. Aside teleportation, other attempts have been made to addressing the lack of ergodicity. \citet{zhou2005learning} build on the directed Laplacian of \citet{chung2005laplacians} to develop a spectral clustering method tailored to digraphs. \citet{meilua2007clustering} directly leverage the asymmetric adjacency matrix within a weighted cut framework. More recently, \citet{rohe2016co} proposed a spectral co-clustering approach based on a singular value decomposition, capturing directional information through low-rank approximations.

Another prominent source of digraphs arises from point-cloud data, where $K$-nearest neighbor ($K$-NN) graphs are commonly used. These graphs are typically asymmetric due to the directional nature of distance- or kernel-based construction methods \citep{kernelsML2001}. In this context, it is common to symmetrize the adjacency matrix and apply spectral clustering on the resulting undirected graph Laplacian \citep{satuluri2011symmetrizations}. While convenient, this transformation can obscure directional patterns and degrade performance.

A rather different line of work focuses on \emph{flow-based clustering} \citep{cucuringu2020hermitian, laenen2020higher, coste2021simpler, hayashi2022skew}, aiming at identifying \emph{imbalanced cuts}, \ie partitions where most edges flow in one direction between clusters. This is particularly suited to domains with strongly asymmetric dynamics, such as migration networks \citep{cucuringu2020hermitian}, food webs, or trade flows \citep{laenen2020higher}. However, they are distinct from the classical \emph{density-based} clustering, which seeks balanced clusters with high internal cohesion.

In this work, we revisit the density-based clustering paradigm and we propose a novel framework that generalizes classical spectral methods to both directed and undirected graphs. Our contributions are as follows:
\begin{itemize}[leftmargin=1.4em, topsep=-0.5em, itemsep=-0.3em]
    \item \textbf{\textit{Generalized Dirichlet energy} (\GDE):} We introduce a %
		framework based on a novel energy functional that extends the classical Dirichlet energy \citep{montenegro2006mathematical}. Unlike prior formulations, which are limited to ergodic random walks with specific transition matrices and stationary distributions, \GDE is defined for \emph{any} positive vertex measure and Markov transition matrix, without requiring reversibility or ergodicity. It arises naturally from a random walk interpretation of graph partitioning. Moreover, we propose a parametrized vertex measure that encodes both directionality and edge density. %

    \item \textbf{\textit{Generalized spectral clustering} (\GSC):} Building on the \GDE, we develop a spectral clustering method that applies to both directed and undirected graphs. Unlike earlier spectral approaches that are limited to strongly connected digraphs \citep{zhou2005learning, palmer2020spectral}, \GSC applies to \emph{weakly connected} digraphs without relying on teleporting random walks, such as those used in PageRank \citep{page1999pagerank}. This leads to a principled and flexible clustering method suited for a wide variety of graph-structured data.
\end{itemize}
\newpage

\section{Preliminaries and background}
\label{sec:background}

\subsection{Essential concepts}
\label{sec:essential-concepts}

\inlinetitle{Notation}{.}%
Let $\mathcal{G} = (\gV, \mathcal{E}, w)$ be a weighted directed graph (digraph), where $\gV$ is a finite vertex set of size $N = |\gV|$, and $\mathcal{E} \subseteq \gV \times \gV$ is a set of directed edges. Each edge $(i, j)$ denotes a directed link from vertex $i$ to vertex $j$. Digraphs may be given directly or constructed from input point-clouds, i.e. from datapoints $\boldsymbol{X} = \{x_i\}_{i=1}^N$, with $x_i \in \mathbb{R}^d$, using appropriate graph construction techniques. We assume that $\mathcal{G}$  is a \emph{weakly connected} digraph: that is, when the edge directionality is ignored, the obtained graph is connected.

The edge weight function $w: \gV \times \gV \to \mathbb{R}_+$ assigns a nonnegative value to each vertex pair: $w(i,j) > 0$ if $(i,j) \in \mathcal{E}$, and $0$ otherwise. Let $\rmW = \{w_{ij}\}_{i,j=1}^N \in \mathbb{R}_+^{N \times N}$ be the adjacency matrix, where $w_{ij} = w(i,j)$. Define the out-degree and in-degree of vertex $i$ as $d_\textnormal{out}(i) = \sum_j w_{ij}$ and $d_\textnormal{in}(i) = \sum_j w_{ji}$, respectively. For a vector $\nu$, define $\rmD_\nu = \diag{\nu}$. For a subset $S \subseteq \gV$, its complement is $\bar{S} = \gV \setminus S$, and its characteristic function is $\chivec{S} \in \{0,1\}^N$, where $\chivec{S}(i) = 1$ iff $i \in S$. For a singleton set $S = \{v\}$, we use $\delta_v \in \{0,1\}^N$, the Kronecker delta vector. Also, $\one_{N \times M}$ is the $N \times M$ all-ones matrix, and $\mathds{1}\{\cdot\}$ is the indicator function.

Any graph function $\nu: \gV \to \mathbb{R}_+$ assigning nonnegative values to vertices can be seen as a \emph{positive vertex measure}, when $\sum_{i \in \gV} \nu(i) = 1$, $\nu$ is a \emph{probability vertex measure}. A function $f : \gV \to \mathbb{R}$ is represented as $f = [f(i)]_{i \in \gV}^\top \in \mathbb{R}^N$. We assume graph functions belong to $\ell^2(\gV, \nu)$, a Hilbert space with inner product:
\begin{equation}
  \langle f, g \rangle_{\nu} = f^\top \rmD_{\nu} g = \sum_{i \in \gV} \nu(i) f(i) g(i),
\end{equation}
which reduces to the standard dot product $\langle f, g \rangle = f^\top g$ when $\nu = \one_{N \times 1}$. Edge-based functions $q: \gV \times \gV \to \mathbb{R}_+$, constrained to $q(i,j) = 0$ for $(i,j) \notin \mathcal{E}$, are called \emph{positive edge measures}.

\inlinetitle{Random walk fundamentals}{.}%
A natural random walk on $\mathcal{G}$ is defined by a homogeneous Markov chain $\mathcal{X} = (X_t)_{t \geq 0}$ on a state space $\gV$, with transition probabilities:
\begin{equation}
  p(i,j) = \mathbb{P}(X_{t+1} = j \mid X_t = i) = \frac{w(i,j)}{\sum_z w(i,z)}.
\end{equation}
The transition matrix $\rmP = [p(i,j)] \in \mathbb{R}^{N \times N}$ is row-stochastic, and equals: $\rmP = \rmD_{\textnormal{out}}^{-1} \rmW$. Its spectrum lies within the unit disk, \ie all its eigenvalues have absolute value at most $1$.
If $\mathcal{G}$ is strongly connected and aperiodic, then the random walk is ergodic, and the distribution $p_t(i, \cdot) = \delta_i^\top \rmP^t$ converges to a unique stationary distribution $\pi \in \mathbb{R}_+^N$ as $t \to \infty$ \citep{bremaud2013markov}.
For undirected graphs or reversible walks, the stationary distribution satisfies $\pi(i) \propto d(i)$, the degree of vertex $i$. Reversibility means $\pi(i) p(i,j) = \pi(j) p(j,i)$ for all $i,j$. This condition always holds in the undirected setting, where $d_{\textnormal{out}}(i) = d_{\textnormal{in}}(i) = d(i)$.

\inlinetitle{Dirichlet energy and graph Laplacians}{.}%
The \textit{Dirichlet energy} quantifies the global smoothness of a graph function, penalizing variation across edges. It is central in Dirichlet form theory \citep{saloff1997lectures, montenegro2006mathematical}, graph signal processing \citep{GSP2012}, and harmonic analysis on graphs \citep{sevi2023harmonic}.
\begin{definition}[\textbf{Dirichlet energy of a graph function}]
Let $\mathcal{X}$ be a random walk on digraph $\mathcal{G}$ with transition matrix $\rmP$ and stationary distribution $\pi$. Then the Dirichlet energy of $f : \gV \to \mathbb{R}$ is:
\begin{align}
  \D{}{f} &= \sum_{i,j \in \gV} \pi(i) p(i,j) [f(i) - f(j)]^2 \, \label{eq:Dirichlet} \\
           &= 2\, \langle f, \rmL_{\textnormal{RW}} f \rangle_{\pi} = 2\, \langle f, \rmL f \rangle. \label{eq:Dirichlet-quadratic}
\end{align}
\end{definition}
Eq.~\eqref{eq:Dirichlet} encourages smooth functions over high-probability transitions. Eq.~\eqref{eq:Dirichlet-quadratic} links the Dirichlet energy to the Laplacian matrices $\rmL_{\textnormal{RW}}$, $\rmL$, $\overline{\rmL}$ \citep{chung2005laplacians,sevi2023harmonic}, which are fundamental to spectral methods.
\begin{definition}[\textbf{Directed Laplacians}]
Given a transition matrix $\rmP$ and its stationary distribution $\pi$, the directed Laplacians are defined as follows:
\begin{align}
  \text{Random walk Laplacian:} & \quad \rmL_{\textnormal{RW}} = \mathbf{I} - \tfrac{1}{2}(\rmP + \rmD_{\pi}^{-1} \rmP^\top \rmD_{\pi}) \, \label{eq:rw_lap_d} \\
  \text{Unnormalized Laplacian:} & \quad \rmL = \rmD_{\pi} - \tfrac{1}{2}(\rmD_{\pi} \rmP + \rmP^\top \rmD_{\pi}) \, \label{eq:unnorm_lap_d} \\
  \text{Normalized Laplacian:} & \quad \overline{\rmL} = \rmD_{\pi}^{-1/2} \rmL \rmD_{\pi}^{-1/2}.
\end{align}
\end{definition}
In the undirected case, $\rmD_{\pi} \propto \rmD_d$ and $\rmP = \rmD_d^{-1} \rmW$, implying:
\begin{equation}
  \rmL_{\textnormal{RW}} = \mathbf{I} - \rmP, \quad \rmL = \rmD_d - \rmW.
\end{equation}
For balanced graphs (equal in- and out-degrees), these Laplacians reduce to those for undirected graphs using the symmetrized matrix $\rmW_{\text{sym}} = \tfrac{1}{2}(\rmW + \rmW^\top)$.

\subsection{Connection with spectral clustering}
\label{sec:connection-to-spectral-clustering}

The Dirichlet energy yields a Laplacian operator, connecting it to spectral methods. In spectral clustering, node embeddings are obtained from the eigenvectors of a chosen Laplacian.
To formalize this connection, we recall the Courant-Fischer min-max theorem \citep{matrixanalysis1985} from spectral theory.
For a Hermitian matrix $\rmL$ with eigenvalues $\lambda_1 \geq \dots \geq \lambda_N$, the Rayleigh quotient
\begin{equation}\label{eq:rayleigh-quotient}
  \mathcal{R}_{\rmL}(y) = \frac{y^\top \rmL y}{y^\top y} = \frac{\langle y, \rmL y \rangle}{\|y\|^2}, \quad y \in \mathbb{R}^N \setminus \{0_N\},
\end{equation}
characterizes the eigenvalues via
\begin{align}
  \lambda_k = \max_{\substack{M \subset \mathbb{R}^N \\ \dim(M) = N-k+1}} \min_{y \in M} \mathcal{R}_{\rmL}(y).
\end{align}
The $k$-th eigenvector, $v_k$, minimizes $\mathcal{R}_{\rmL}$, subject to orthogonality:
\begin{align}\label{eq:rayleigh-ritz}
\lambda_k = \min_{y \in \mathbb{R}^N} \mathcal{R}_{\rmL}(y), \quad
v_k = \argmin_{y \in \mathbb{R}^N} \mathcal{R}_{\rmL}(y) \\
\text{subject to } \langle v_j, y \rangle = 0, \text{ for } j=1,\dots,k-1.
\end{align}
\begin{remark}\label{remark:Dirichlet-Rayleigh-equivalence}
The Dirichlet energy and the Rayleigh quotient are proportional:
\begin{align}
  \Dnorm{}{f} = 2 \, \mathcal{R}_{\rmL}(f), \quad \Rightarrow \quad \min_{f \in \R^N} \mathcal{R}_{\rmL}(f) \equiv \min_{f \in \R^N} \Dnorm{}{f}, \label{eq:Dirichlet-Rayleigh-equivalence}
\end{align}
where $\Dnorm{}{f} = \frac{\D{}{f}}{\|f\|^2}$ is the \emph{normalized Dirichlet energy}.
\end{remark}
Therefore, minimizing the Rayleigh quotient is equivalent to minimizing Dirichlet energy. This also reveals that minimizing the Dirichlet energy promotes smooth functions over a graph, assigning similar values to well-connected nodes. For binary clustering ($k=2$), one may use $f = \chivec{V_1}$, the indicator of cluster $V_1$. This generalizes naturally to a multi-way partitioning (see Sec.~\ref{sec:gsc}).

\section{The generalized Dirichlet energy framework} \label{sect:gen_dir_en_L}

\subsection{Generalized Dirichlet energy and graph Laplacians}

We now formalize the connection between Dirichlet energy and graph Laplacians through the introduction of the \emph{generalized Dirichlet energy} (\GDE). This functional is defined with respect to a general positive edge measure $\q$, and naturally leads to a broader class of graph Laplacians. We begin with the general definition of \GDE, then specialize to cases where $\q$ is derived from a vertex measure $\nu$ and a transition matrix $\rmP$. This leads to a family of Laplacian operators that adapt to the graph structure and the random walk dynamics. Although our focus is spectral clustering, the operators defined through the \GDE framework can be of use in other graph-based learning problems.
\begin{definition}[\textbf{Generalized Dirichlet Energy of a graph function}] \label{def:gen_dir_energy}
  Let $\q : \gV \times \gV \to \mathbb{R}_+$ be a positive edge measure on a digraph $\mathcal{G}$, and let $\Q = [\q(i,j)]_{i,j \in \gV}$ denote the corresponding matrix. The \emph{generalized Dirichlet energy} of a function $f : \gV \to \mathbb{R}$ is given by:
  \begin{equation} \label{full_gen_dir_nrj}
      \D{\Q}{f} = \sum_{i,j \in \gV} \q(i,j) [f(i) - f(j)]^2.
  \end{equation}
\end{definition}
This formulation is flexible: the graph structure is captured by the choice of edge measure $\q$. For instance, setting $\q(i,j) = \pi(i)p(i,j)$ where $\pi$ is the stationary distribution of a random walk with transition matrix $\rmP$, is a special case yielding the classical Dirichlet energy:
\begin{equation}
  \D{\rmD_\pi \rmP}{f} = \sum_{i,j \in \gV} \pi(i) p(i,j) [f(i) - f(j)]^2 =  \D{}{f}.
\end{equation}
More generally, we adopt the factorized form $\q(i,j) = \nu(i)p(i,j)$, with $\nu$ being a tunable vertex measure. With this we define:
\begin{equation} \label{eq:dir_en_P}
  \D{\nu}{f} = \sum_{i,j \in \gV} \nu(i)p(i,j)[f(i) - f(j)]^2,
\end{equation}
which enables new energy functionals %
with arbitrary positive vertex measures $\nu$.
Since scaling $\nu$ simply rescales the energy by $\norm{\nu}_1^{-1}$, we may assume without loss of generality that $\nu$ is a probability vertex measure.
\begin{definition}[\textbf{Generalized graph Laplacians}] \label{def:gen_lap}
  Let $\mathcal{X}$ be a random walk on a digraph $\mathcal{G}$ with transition matrix $\rmP$. Let $\nu$ be a positive vertex measure, and define the incoming measure $\xi = \rmP^\top \nu$. Define $\rmD_{\nu+\xi} = \rmD_\nu + \rmD_\xi$. The associated generalized Laplacians are:
  \begin{align}
      \text{Unnormalized:} &\quad \rmL_\nu = \rmD_{\nu+\xi} - (\rmD_\nu \rmP + \rmP^\top \rmD_\nu), \label{unnorm_glap} \\
      \text{Random walk:} &\quad \rmL_{\textnormal{RW},\nu} = \rmD_{\nu+\xi}^{-1} \rmL_\nu, \label{gen_rw_lap} \\
      \text{Normalized:} &\quad \overline{\rmL}_\nu = \rmD_{\nu+\xi}^{-1/2} \rmL_\nu \rmD_{\nu+\xi}^{-1/2}.
  \end{align}
\end{definition}
Here, $\xi = \rmP^\top \nu$ corresponds to the Perron--Frobenius operator applied to $\nu$ \citep{Ding2009,klus2023transfer,klus2023koopman}. When $\nu = \pi$, %
we recover classical Laplacians:
\begin{equation}
  \rmL_\pi = 2\rmD_\pi - (\rmD_\pi \rmP + \rmP^\top \rmD_\pi) = 2\rmL.
\end{equation}
\vspace{-1.5em}
\begin{restatable}[]{property}{gdeglap}
  \label{gde_glap}
  Let $\nu$ be a positive vertex measure and $\xi := \rmP^\top \nu$. Then for any graph function $f$:
  \begin{equation}
    \D{\nu}{f} = \langle f, \rmL_{\textnormal{RW},\nu} f \rangle_{\nu+\xi} = \langle f, \rmL_\nu f \rangle.
  \end{equation}
\end{restatable}
This implies that $\rmL_{\textnormal{RW},\nu}$ is self-adjoint in $\ell^2(\gV, \nu + \xi)$, while $\rmL_\nu$ and $\overline{\rmL}_\nu$ are self-adjoint in $\ell^2(\gV, \nu)$.

\inlinetitle{Rayleigh quotient}{.}%
The normalized generalized Dirichlet energy is given by:
\begin{equation}
  \label{rayleigh_ratio}
  \Dnorm{\nu}{f} = \frac{\D{\nu}{f}}{\|f\|_{\nu + \xi}^2}, \quad \text{where } \|f\|_{\nu + \xi}^2 = \langle f, \rmD_{\nu + \xi} f \rangle.
\end{equation}
This aligns with the Rayleigh quotient principle:
\begin{equation}
  \mathcal{R}_{\rmL_\nu}(f) = \frac{\langle f, \rmL_\nu f \rangle}{\langle f, \rmD_{\nu + \xi} f \rangle}.
\end{equation}

\subsection{Parametrized vertex measure} \label{sec:parametrized-vertex-measure}

We now construct vertex measures, denoted by $\nu$, that capture local and global graph dynamics via random walks. The underlying idea is that, even when the stationary distribution $\pi$ exists, other parametrizations can capture better the graph geometry.
Starting from a uniform distribution, we define the following family of measures that depend on the diffusion time $t$:
\begin{equation} \label{eq:basic_nu}
  \nu_t = [\rmP^t]^\top \tfrac{1}{N} \one_{N \times 1}.
\end{equation}

\begin{restatable}[]{property}{dirichlett}
\label{prop:dirichlet_t}
Let $\nu_t = [\rmP^t]^\top \tfrac{1}{N} \one_{N \times 1}$, where $\rmP$ is ergodic. Then:
\begin{equation}
  \lim_{t \to \infty} \D{\nu_t}{f} = \D{\pi}{f} = \D{}{f}.
\end{equation}
\end{restatable}
Note that the above property holds for any probability measure $\nu = [\rmP^t]^\top \mu$, where $\mu$ is a probability vertex measure; in Eq.~\ref{eq:basic_nu}, though, $\mu$ is taken to be the uniform measure. Moreover, for any time-based vertex measure $\nu_t$ such that $\nu_t \xrightarrow{t \to \infty} \nu$, it holds that $\D{\nu_t}{f} \xrightarrow{t \to \infty} \D{\nu}{f}$.

On the top of the base measure $\nu_t$, we can define the following more flexible parametrized measure: \\
Given $t \in \mathbb{N}$ and $\alpha \in \mathbb{R}_+$,
\begin{equation}\label{eq:refined_nu}
  \nu_{t,\alpha} = \left( [\rmP^t]^\top \tfrac{1}{N} \one_{N \times 1} \right)^{\odot \alpha}, \quad \text{i.e. } \nu_{t,\alpha}(i) = \left( [\rmP^t]^\top \tfrac{1}{N} \one_{N \times 1} \right)_i^\alpha.
\end{equation}

The proposed family of vertex measures $\nu_{t, \alpha}$ is grounded in two key principles. First, the distribution $\nu_{t, \cdot}$ reflects the outcome of a $t$-step diffusion starting from a uniform distribution. For reversible and ergodic random walks, as $t \to \infty$, $\nu_{t, \alpha}$ converges to the stationary distribution $\pi$, enabling a smooth interpolation between the local and the global view of a graph. Second, the exponent $\alpha$ acts as a sharpness parameter, inspired by transformations used in normalized Laplacians and information-theoretic reweightings. Varying $\alpha$ allows the control of the distribution's concentration: $\alpha < 1$ flattens the measure, $\alpha > 1$ sharpens it, and $\alpha = 0$ makes the measure uniform. This formulation offers greater flexibility to adapt to graph heterogeneity. Note that, setting $\alpha \neq 1$ gives a non-probability measure, which however does not affect the essence or validity of the model. %
Plugging the proposed $\nu_{t,\alpha}$ into Eq.~\eqref{eq:dir_en_P} yields:
\begin{equation} \label{gde:power_alpha}
  \D{\nu_{t,\alpha}}{f} = \langle f, \rmL_{t,\alpha} f \rangle,
\end{equation}
where $\rmL_{t,\alpha}$ is the generalized Laplacian associated with the reweighted measure $\nu_{t,\alpha}$. The proposed $\nu_{t,\alpha}$ %
enables smooth transitions between local/global and sharp/flat configurations via the computation of a family of operators $(\rmL_{t, \alpha})_{t, \alpha}$. This family of operators supports adaptive, data-driven spectral clustering, with a high focus on flexibility.

\section{Generalized spectral clustering via GDE minimization} \label{sec:gsc}
The \emph{generalized spectral clustering} (GSC) method, based on \GDE minimization, is applicable to any graph, directed or not. The method stems from the random walk viewpoint of graph partitioning. We start by revisiting the random walk perspective of graph $2$-partitioning and we then pass to the $k$-partitioning.

\subsection{\GDE of a graph \texorpdfstring{$\boldsymbol{2}$}{2}-partition}
\label{sec:2-way-GDE}
For an arbitrary probability vertex measure $\nu: \gV \mapsto \mathbb{R}_{+}$, let $\nu(S)$ be its evaluation over a subset $S \subseteq \gV $: $\nu(S) = \sum_{i \in S} \nu(i)$. Let also $\q:\mathcal{E}\mapsto \mathbb{R}_{+}$ be the composite edge measure such that $\q(i,j) = \nu(i)p(i,j)$. Respectively, we can define $\q(S,U)$ to be the edge measure between two disjoint vertex subsets $S,U\subseteq \gV $, $S \cap U = \emptyset$, and generally $S \cup U \neq \gV $:
\begin{align}
  \q(S,U)&=\sum_{i\in S, j\in U}\q(i,j)\ = \sum_{i\in S, j\in U} \nu(i)p(i,j), \nonumber\\
  &= \sum_{i\in S, j\in U}\mathbb{P}(X_t=i) \mathbb{P}(X_{t+1}=j\padded{|}X_t=i),\nonumber\\
  &=\mathbb{P}(X_t\in S,X_{t+1}\in U), \ \ \text{for any } t\geq 0. \label{eq:rw_cut}
\end{align}
The generic measure $\q(S,U)$ is related to an ergodic Markov chain at the equilibrium, \ie when $\nu = \pi$ \citep{sinclair1992improved,levin2017markov}. In our setting, we generalize $\q(S,U)$ for any transition matrix and any vertex measure, such that it quantifies the probability that the random walk escapes from the set $S$ to $U$ in one step, when the starting vertex of the walk is drawn according to the arbitrary vertex measure $\nu$.
When considering $U = \bar{S}$, this discussion becomes very interesting for graph partitioning. In essence, $\q(S,\bar{S})$ offers a probabilistic point of view over the graph cut between the set $S$ and the rest of the graph.

Note that \cite{meilua2001random} have presented a similar measure %
in the ergodic setting $\nu = \pi$. Next, we establish the connection between \GDE and the edge measure $\q(S,\Bar{S)}$ (the proof is in Appendix\,\ref{proof:btlgde}).

\begin{restatable}[]{property}{btlgde}\label{prop:btl_gde}
  Let $\mathcal{X}$ be a random walk on a digraph $\mathcal{G}$, with transition matrix $\rmP$. Let $\nu$ be a positive vertex measure, and $\q$ be its associated positive edge measure, both on $\mathcal{G}$. Let $S \subseteq \gV $ and $\Bar{S} = \gV  \backslash S$. Consider the characteristic function $\chivec{S}$, associated with the set $S$, as a graph function. The composite edge measure $\q(S,\Bar{S})$ and the \GDE are related as follows:
  \begin{equation}\label{eq:dir_decision_prop}
    \q(S,\Bar{S})+\q(\Bar{S},S)=\D{\nu}{\chivec{S}}.
  \end{equation}
\end{restatable}
In the case of a $2$-partitioning, the function $\chivec{S}$ serves as an indicator (partition) function that separates the vertex set into two disjoint parts, $S$ and $\bar{S}$ (see also Sec.\,\ref{sec:connection-to-spectral-clustering}). Eq.~\eqref{eq:dir_decision_prop} thus provides a natural interpretation of the generalized Dirichlet energy for graph partitions: the quantity $\D{\nu}{\chivec{S}}$ measures how easily a random walk with transition probabilities $\q$ escapes from $S$ to $\bar{S}$, or vice versa. This energy is symmetric by construction, since $\D{\nu}{\chivec{S}} = \D{\nu}{\chivec{\bar{S}}}$. Under mild assumptions\footnote{The graph defined by the transition matrix $\q(i,j)$ must be connected and aperiodic (non-bipartite).}, one can further show that $\D{\nu}{\chivec{S}}$ corresponds to the normalized cut (N-Cut) of a graph with edge weights $\q(i,j)$. In this context, the vertex measure $\nu$ plays the role of a regularizer in the graph cut objective, shaping the structure of the optimal partition.

When the vertex measure is chosen as the stationary distribution $\nu = \pi$ of the random walk, the following corollary holds (the proof is in Appendix~\ref{proof:btl_gde}).

\begin{corollary}[of Proposition~\ref{prop:btl_gde}]
  Let $\mathcal{X}$ be an ergodic random walk on a digraph $\mathcal{G}$, with transition matrix $\rmP$ and ergodic (stationary) distribution $\pi$. For any subset $S \subseteq \gV$, let $\Bar{S} = \gV \setminus S$, and let $\chivec{S}$ be the indicator function of $S$. Then, the composite edge measure $\q(S,\Bar{S})$ associated with $\pi$ and the generalized Dirichlet energy are related as:
  \begin{equation} \label{corr:prop:btl_gde}
    \q(S,\Bar{S}) = \frac{1}{2} \D{\pi}{\chivec{S}}.
  \end{equation}
\end{corollary}

This corollary reveals a key insight: minimizing the Dirichlet energy $\D{\pi}{\chivec{S}}$ is equivalent to minimizing the total ergodic flow crossing between $S$ and its complement $\Bar{S}$. In other words, $\D{\pi}{\chivec{S}}$ serves as a normalized cut-like objective under the stationary distribution of the random walk. More generally, in this work we establish the idea that if a probability measure $\nu$ is used in the place of $\pi$ (\eg a measure that approximates or regularizes $\pi$) then minimizing $\D{\nu}{\chivec{S}}$ can still be interpreted as finding a well-separated $2$-partition of the graph under the dynamics induced by $\nu$.

\subsection{\GDE minimization for graph \texorpdfstring{$\boldsymbol{k}$}{k}-partitioning}
\label{sec:k-way-GDE}
The goal of graph $k$-partitioning is to divide the vertex set of a digraph into $k$ disjoint subsets such that the edge density across subsets is minimized. Let $\boldsymbol{V} = \{V_\kappa\}_{\kappa=1}^k$ denote such a partition, where each $V_\kappa \subset \gV$ is a subset of the vertex set of the graph, and $V_\kappa \cap V_{\kappa'} = \emptyset$ for $\kappa \neq \kappa'$, with $\bigcup_{\kappa=1}^k V_\kappa = \gV$.
Associated with each subset $V_\kappa$, there is a partition function $\chivec{V_\kappa} \in \{0,1\}^N$ that serves as an indicator vector over the vertices. The partition functions define a set of binary signals $\{ \chivec{V_\kappa} \}_{\kappa=1}^k$, each highlighting one cluster versus the rest.

We define the \emph{Dirichlet energy of the $k$-partition} under a vertex measure $\nu$ as:%
\begin{equation}\label{eq:dir_k-partition}
  \D{\nu}{\boldsymbol{V}} = \sum_{\kappa=1}^{k} \D{\nu}{\chivec{V_\kappa}},
\end{equation}%
where each term measures the generalized Dirichlet energy associated with a single cluster.

Let $\rmU = [u_1 \ u_2 \ \cdots \ u_k] \in \mathbb{R}^{N \times k}$ be the matrix whose columns are the indicator vectors $u_\kappa = \chivec{V_\kappa}$. The generalized spectral clustering (GSC) method, based on the \GDE framework, casts the graph partitioning task as the following optimization problem:
\begin{equation}\label{eq:gen_sc_1}
  \min_{\boldsymbol{V} = \{V_1, \dots, V_k\}} \ \D{\nu}{\boldsymbol{V}}
  = \min_{\rmU} \ \textnormal{tr}(\rmU^\top \rmL_\nu \rmU)
  \quad \textnormal{s.t.} \ \ u_\kappa = \chivec{V_\kappa},\ \forall \kappa \in \{1,\dots,k\}.
\end{equation}
As with classical spectral clustering, this problem is \textnormal{NP}-hard due to the discrete constraints on $\rmU$ \citep{von2007tutorial}. To make it tractable, we adopt the standard relaxation: instead of binary indicator functions, we allow $\rmU$ to be any real-valued matrix with orthonormal columns. This leads to the relaxed optimization problem:
\begin{equation}\label{gen_sc}
  \min_{\rmU} \ \textnormal{tr}(\rmU^\top \rmL_\nu \rmU)
  \quad \textnormal{s.t.} \ \ \rmU^\top \rmU = \mathbf{I}_k,
\end{equation}
whose solution is given by the $k$ eigenvectors of $\rmL_\nu$ corresponding to its smallest eigenvalues.

The novelty of the proposed method lies in the generalized Laplacian $\rmL_\nu$, which is defined with respect to an arbitrary positive vertex measure $\nu$. This flexibility stems from the \GDE framework and captures a rich family of partitioning behaviors grounded in random walk dynamics. When the measure is chosen as the stationary distribution, $\nu = \pi$, and the normalized Dirichlet energy (see Eq.\,\eqref{rayleigh_ratio}) is minimized, the resulting method recovers the classical approach for strongly connected digraphs. \citep{zhou2005learning}.

\inlinetitle{Clustering algorithm}{.}%
The \GSC framework relies on a parametrized vertex measure $\nu := \nu_{t,\alpha}$ introduced in Sec.~\ref{sec:parametrized-vertex-measure}. Given a pair of parameter values $(t \in \mathbb{N}, \,\alpha \in \mathbb{R}_+)$, Eq.~\ref{eq:refined_nu} defines the vertex measure $\nu_{t,\alpha}$ that induces a generalized Dirichlet energy $\D{\nu_{t,\alpha}}{\boldsymbol{V}}$ (Eq.~\ref{eq:dir_k-partition}). This, in turn, leads to the generalized spectral graph partitioning objective of Eq.~\ref{gen_sc}, based on the generalized Laplacian $\rmL_{t,\alpha}$ (see Def.~\ref{gen_rw_lap}).
In practice, the \GSC algorithm follows the same computational pipeline of the classical spectral clustering (see Alg.~\ref{gsc_algo}), with the difference that it substitutes the standard Laplacian with the generalized one $\rmL_{t,\alpha}$, and computes its leading eigenvectors $\rmU_{t,\alpha} \in \mathbb{R}^{N \times k}$ for downstream clustering.

\begin{algorithm}[tb]\small
  \caption{Generalized Spectral Clustering (\GSC)}
  \label{gsc_algo}
  \textbf{Input:} Weighted adjacency matrix $\rmW$; number of clusters $k$; diffusion time $t$; reweighting exponent $\alpha$.\\
  \textbf{Output:} $\boldsymbol{V}_{t,\alpha}$: graph partition into $k$ clusters.
  \vspace{0.5mm}
  \hrule
  \begin{algorithmic}[1]
    \STATE Compute the generalized Laplacian $\rmL_{t,\alpha}$ (Eq.~\ref{unnorm_glap}).
    \STATE Compute $\rmU_{t,\alpha} \in \mathbb{R}^{N \times k}$, whose columns are the eigenvectors corresponding to the $k$ smallest eigenvalues of $\rmL_{t,\alpha}$.
    \STATE Embed each vertex $i$ into $\mathbb{R}^k$ using the $i$-th row of $\rmU_{t,\alpha}$, and apply a clustering algorithm %
		to the embedded points.
    \STATE Construct the vertex partition $\boldsymbol{V}_{t,\alpha} = \{V^{(\kappa)}_{t,\alpha}\}_{\kappa=1}^k$ based on clustering results.
    \STATE \textbf{return} $\boldsymbol{V}_{t,\alpha}$
  \end{algorithmic}
\end{algorithm}

\section{Experiments}
\label{exps_benchmark}

\subsection{Experimental setup}

The proposed \GSC method is empirically evaluated on clustering digraphs constructed from benchmark point-cloud data. The performance is compared against established spectral clustering baselines, under both fully unsupervised and label-informed evaluation protocols.

\inlinetitle{Compared methods}{.}%
We evaluate two \GSC variants based on generalized Laplacians from Sec.~\ref{sec:parametrized-vertex-measure}:
\begin{itemize}[leftmargin=1.4em, topsep=-0.2em, itemsep=-0.5em]
  \item \GSCun($t,\alpha$): using the unnormalized generalized Laplacian $\rmL_\nu$ (Eq.~\ref{unnorm_glap}),
  \item \GSCn($t,\alpha$): using the normalized generalized Laplacian $\overline{\rmL}_\nu$ (Eq.~\ref{gen_rw_lap}),
\end{itemize}
with $\nu := \nu_{t, \alpha}$ being the parametrized vertex measure (Eq.~\ref{eq:refined_nu}). The grid search for tuning the hyperparameters is over the ranges:
$t \in \{0, 1, \dots, 25\}$ and $\alpha \in \{0.0, 0.1, \dots, 1.5\}$, and the upper limits are both set by taking into account the scale of the graph sizes we deal with in the experiments.
We compare against the following spectral clustering methods from the literature, the first four of which have been developed for directed graphs:
\begin{itemize}[leftmargin=1.4em, topsep=-0.2em, itemsep=-0.5em]
  \item DSC$+ (\gamma)$ \citep{zhou2005learning, palmer2020spectral}, extended via teleporting random walks \citep{page1999pagerank} with teleportation parameter $\gamma$.
  \item DI-SIM$_\text{L}(\tau)$, DI-SIM$_\text{R}(\tau)$, and DI-SIM$_\text{C}(\tau)$ \citep{rohe2016co}, which apply spectral clustering to the left, right, or concatenated singular vectors of a regularized transition matrix. For a graph with $n$ nodes and average degree $\bar{d}$:  $\tau \in \left\{ \text{round}\left( \bar{d} \cdot 10^{s} \right) \,\middle|\, s \in \{-1, -0.5, \dots, 1\} \right\}$.
  \item SC$_{\text{un}}$ and SC$_{\text{n}}$: spectral clustering using unnormalized and normalized Laplacians on the symmetrized adjacency matrix \citep{von2007tutorial}.
\end{itemize}

All methods follow a standard spectral clustering pipeline: a graph embedding is computed via the eigenvectors of a Laplacian-type operator, followed by $k$-means\texttt{++} ($100$ restarts) clustering \citep{kmeans++2007} on the rows of the embedding, where the number of clusters $k$ is assumed to be known. The best result is retained, according to on an internal evaluation index.

\inlinetitle{Clustering evaluation protocol}{.}%
The clustering quality is assessed via the following metrics:
\begin{itemize}[leftmargin=1.4em, topsep=-0.2em, itemsep=-0.5em]
  \item \emph{Internal evaluation}: We use an internal evaluation index, the Calinski-Harabasz index (CH), that computes the ratio between the sum of between-cluster variance and the within-cluster variance, for which it utilizes cluster centers. Higher CH values are better.
	\item \emph{External evaluation}: Adjusted Mutual Information (AMI) \citep{strehl2002cluster} is an external evaluation measure computed post hoc and assesses the statistical agreement (adjusted for agreement by chance) between the found clusters and the ground-truth labels. Higher AMI values indicate better clusterings.
\end{itemize}

A \emph{fully unsupervised clustering evaluation protocol} is employed, which emulates real-world conditions where no data labeling is available during training or parameter selection:
\begin{itemize}[leftmargin=1.4em, topsep=-0.2em, itemsep=-0.3em]
\item \emph{Model selection via internal validation}: Each method explores its respective hyperparameter space: \GSC sweeps over $(t, \alpha)$, DSC+ over $\gamma$, and DI-SIM over $\tau$. Each method performs $100$ $k$-means$\texttt{++}$ runs per configuration of its hyperparameters, and the CH index is computed for each clustering solution.
By design, CH prefers more `regular' cluster shapes, meaning spherical (convex and compact) density shapes.
We present experimental results that focus on the clustering methods %
by selecting to use mostly datasets in which CH is suitable.
A workaround to this can be to make a more informed choice of the internal evaluation index and use with the same evaluation pipeline, or to go further and find ways to combine multiple evaluation indices.

\item \emph{Evaluation and reporting}: The solution yielding the best internal score (CH), measured with respect to the the original data representation space, is retained along with the associated optimal hyperparameters; \eg for \GSC, those would be $(t,\alpha)^*_{\textsc{\tiny CH}} := (t^*_{\textsc{\tiny CH}},\alpha^*_{\textsc{\tiny CH}})$. The AMI of that best solution is reported post hoc to interpret better the results using an external evaluation index.
\end{itemize}

\inlinetitle{Graph construction from point-clouds}{.}%
We test the clustering performance across %
real-world datasets from the UCI repository \citep{dheeru2017uci}. Given a dataset $\boldsymbol{X} = \{x_i\}_{i=1}^N$ with $x_i \in \mathbb{R}^d$, we build a sparse directed graph using an unweighted, directed $K$-nearest neighbor ($K$NN) construction, where $K = \lceil \log(N) \rceil$. This process yields sparse and typically weakly connected digraphs.
The unweighted and asymmetric adjacency matrix $\rmW = \{w_{ij}\}$ is defined as: %
$
w_{ij} = \mathds{1}\big\{ \frac{\|x_i - x_j\|^2}{\text{dist}_K(x_i)^2} \leq 1 \big\}
$%
, where $\text{dist}_K (x_i)$ denotes the distance of $x_i$ to its $K$-th nearest neighbor.

\subsection{Results}

\inlinetitle{Sensitivity analysis for the parametrized vertex measure}{.}%
The objective of the first step in our empirical evaluation is two-fold: we demonstrate how the quality of \GSC clustering depends on the vertex measure, we also validate that an internal index such as CH can be used as a surrogate index for unsupervised model selection, through hyperparameter tuning. Fig.~\ref{fig:heatmaps_sensitivity} illustrates the sensitivity of the vertex measure $\nu_{t,\alpha}$ (Eq.~\ref{eq:refined_nu}) to its hyperparameters. For each dataset, two heatmaps are shown with the landscapes produced for the CH and the AMI indices. Shades of orange indicate clustering results of higher quality. A white star in each plot indicates the hyperparameter configuration that yields the best respective evaluation index. Although some variability arises from the $k$-means$\texttt{++}$ initialization, for those datasets poor-quality regions can be identified in the CH landscape, which aligns well with the AMI landscape in high-quality regions.
We can also see that for datasets with well-separated clusters that are easier to detect, the clustering quality is relatively insensitive to the choice of $\nu_{t,\alpha}$; often the measure relies on $\alpha$ or $t$ being equal to $0$. Contrary, in more challenging cases, the choice of measure becomes crucial, and identifying the best hyperparameter becomes a task of interest. These results support using a suitable internal index as a reliable proxy for guiding model selection in a fully unsupervised settings.

\inlinetitle{Model selection and internal evaluation}{.}%
We employ a CH-based model selection to provide a fully unsupervised comparison of the different clustering algorithms. Numerical results are displayed in Tab.~\ref{tab:ch_clustering_scores}, which reports the best CH index of the clustering solutions found by all the compared methods. This is a purely unsupervised evaluation of the clustering quality: in the absence of label information, it is a protocol a user could use to choose the optimal hyperparameters for each method, and the overall `best' solution, by relying to an internal index of choice. The average ranking of the methods across all datasets is also reported, where lower ranks indicate better performance. The competitive index quantifies how close a method is to the best performing method in each dataset, \ie a value closer to $1$ means that they are on average closer the best solutions.
The results show that \GSCn can be efficiently guided to maximize a chosen internal measure, as it achieves the best average rank ($1.5$) in terms of CH. This highlights the strength of the normalized generalized Laplacian paired with the flexible vertex measure $\nu_{t,\alpha}$. \GSCun and SC$_{\text{un}}$ both follow at an average rank of $3.30$. SC$_{\text{n}}$ and DI-SIM$_{\text{L}}$ follow with average ranks around $3.9$ and $4$, respectively. DI-SIM$_{\text{C}}$ follows closely with an average rank of $4.3$. In contrast, DSC$+$ and DI-SIM$_\textsc{R}$ tend to underperform, and are ranked on average lower than $5$-th. The performance gap between \GSCn (rank $1.5$) and \GSCun (rank $3.3$) suggests that the normalized formulation allows for capturing better the data, in particular that the information in the proposed vertex measure might be better highlighted using the normalized variant of \GSC.

\newcommand{\heatmappair}[2]{
    \begin{subfigure}{0.317\textwidth}
      \vspace{6pt}
			\makebox[\linewidth]{\tiny \phantom{xxxxxxx} CH \phantom{xxxxxxxxxxxxxx} AMI \phantom{xxxxxxx}}
			\vspace{-17pt}
			\begin{center}
      \includegraphics[width=0.5\linewidth]{#1_GSC-N_t_alpha_ch.pdf}%
			\includegraphics[width=0.5\linewidth]{#1_GSC-N_t_alpha_ami.pdf}%
			\vspace{-3pt}
			\caption{\footnotesize #2}
			\end{center}
    \end{subfigure}
  }

\begin{figure}[t]
  \centering
  \captionsetup[subfigure]{labelformat=parens, justification=centering}
  \heatmappair{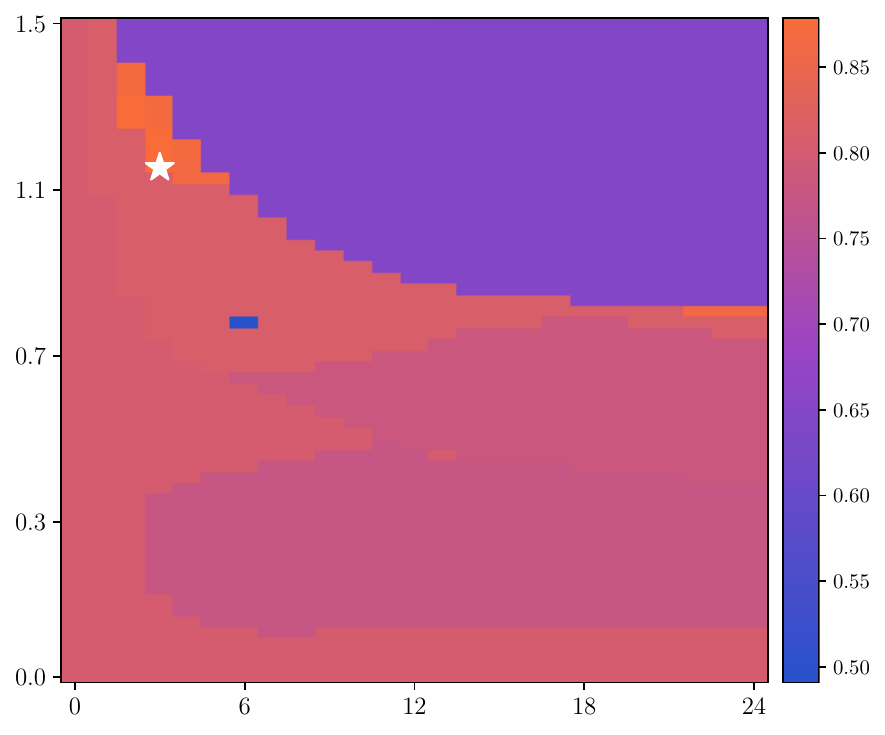}{Iris}
	\heatmappair{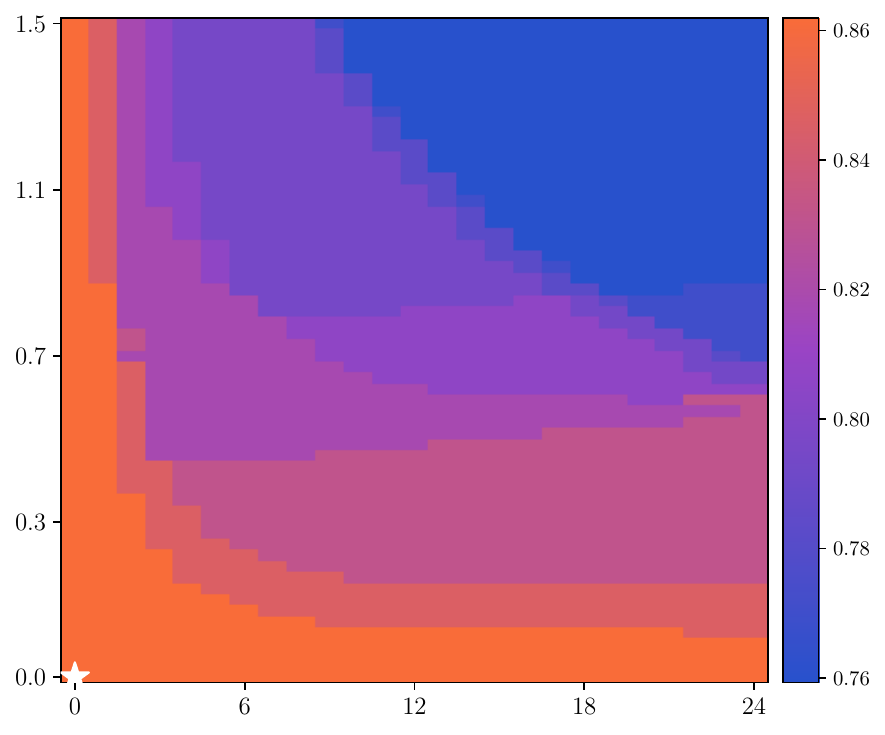}{Wine}
	\heatmappair{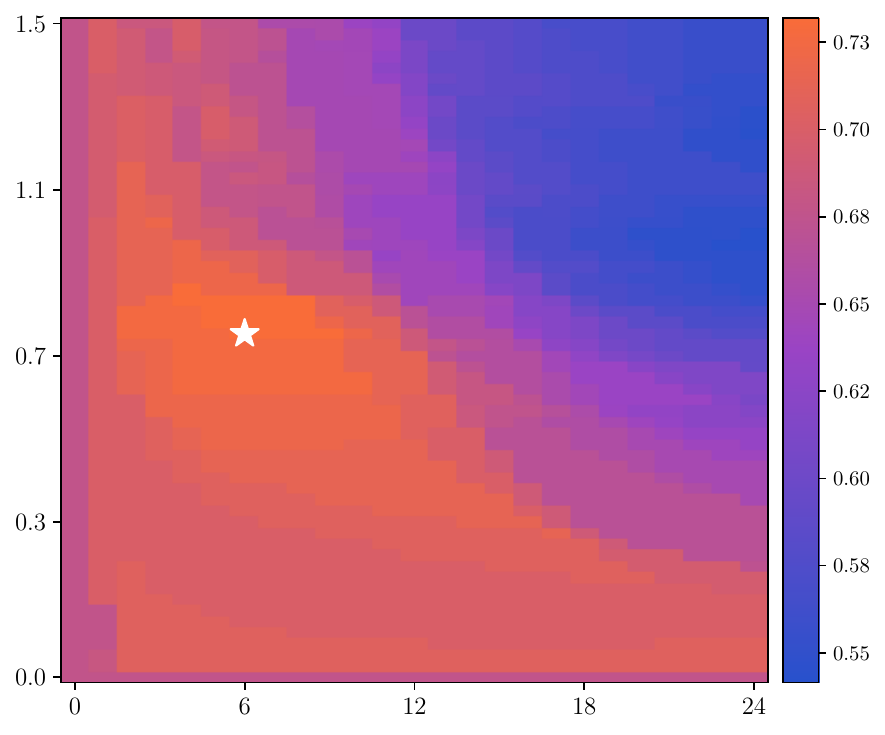}{Wdbc}

  \heatmappair{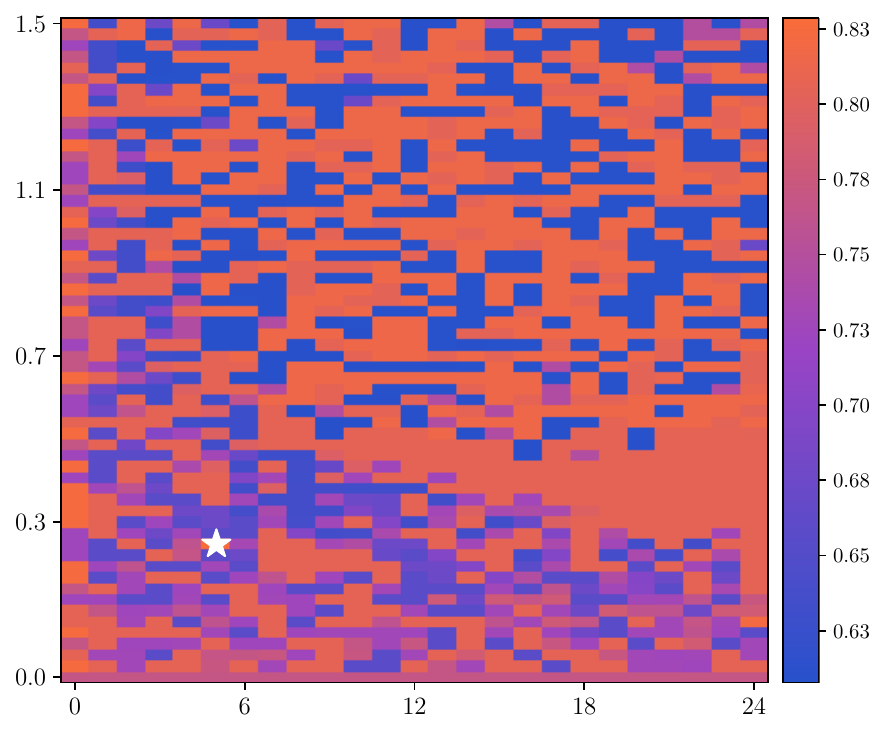}{Control chart}
	\heatmappair{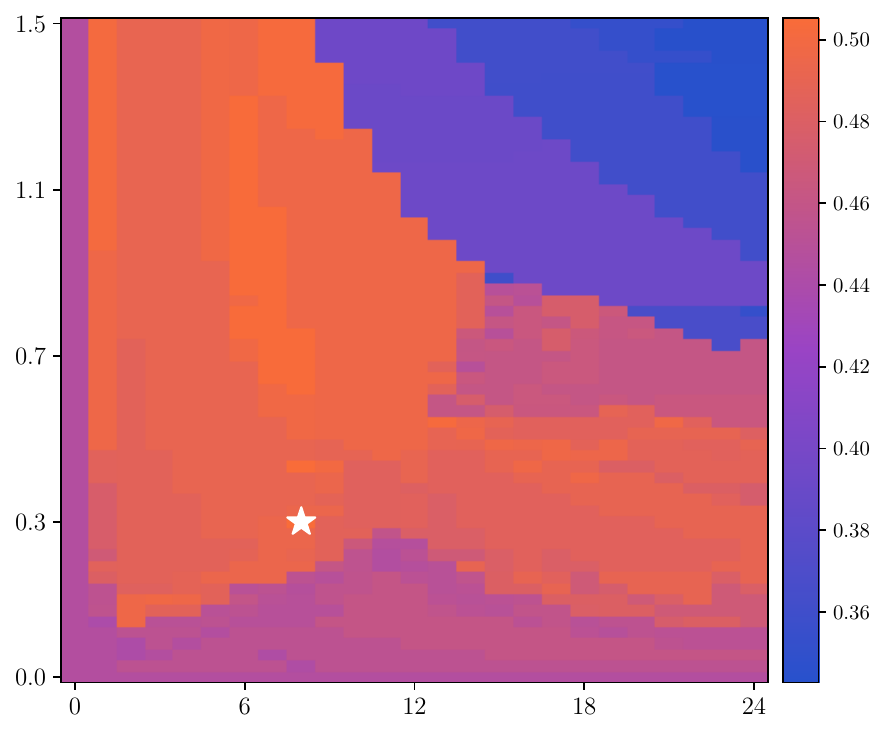}{Breast tissue}
  \heatmappair{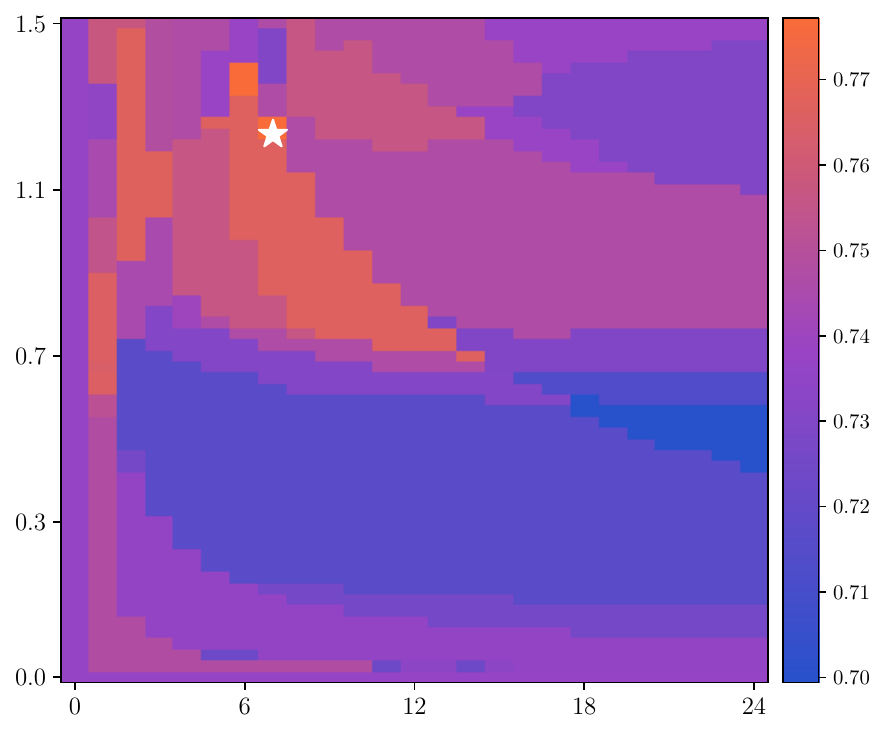}{Seeds}

  \heatmappair{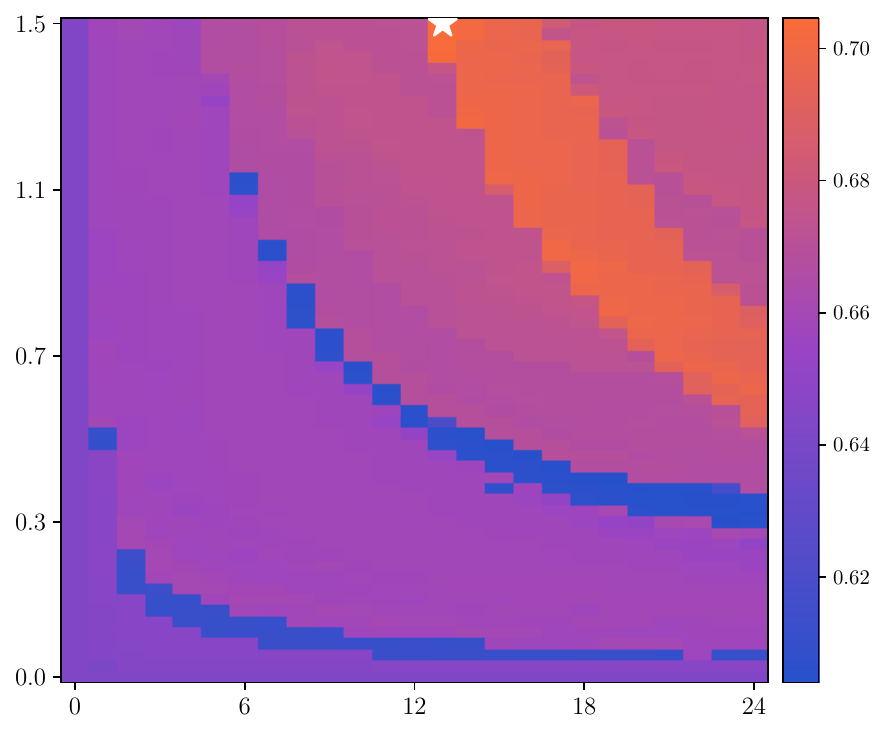}{Segmentation}
  \heatmappair{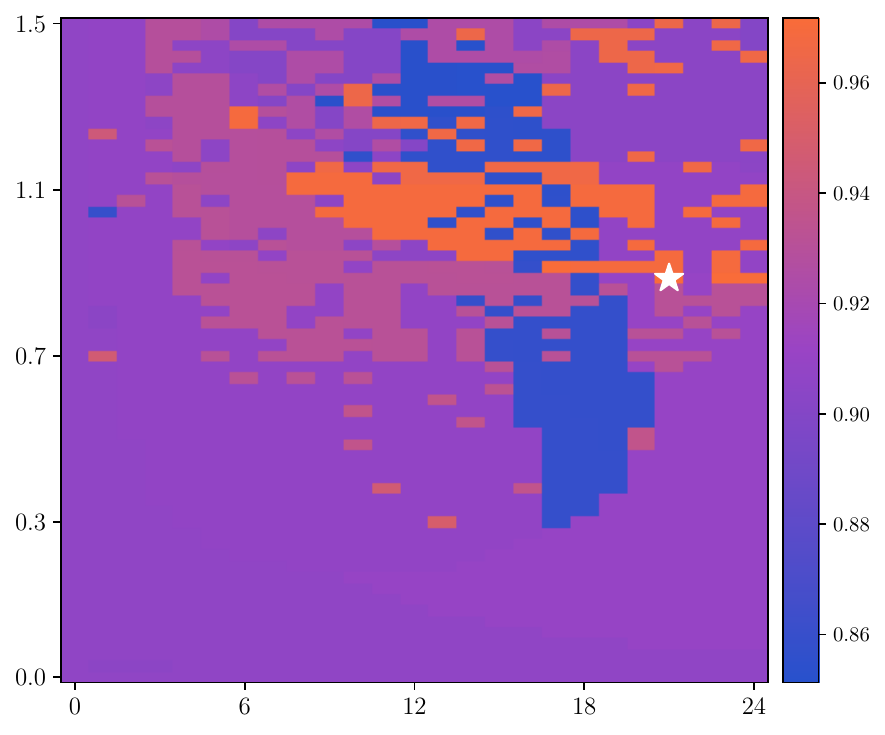}{MNIST64}
  \heatmappair{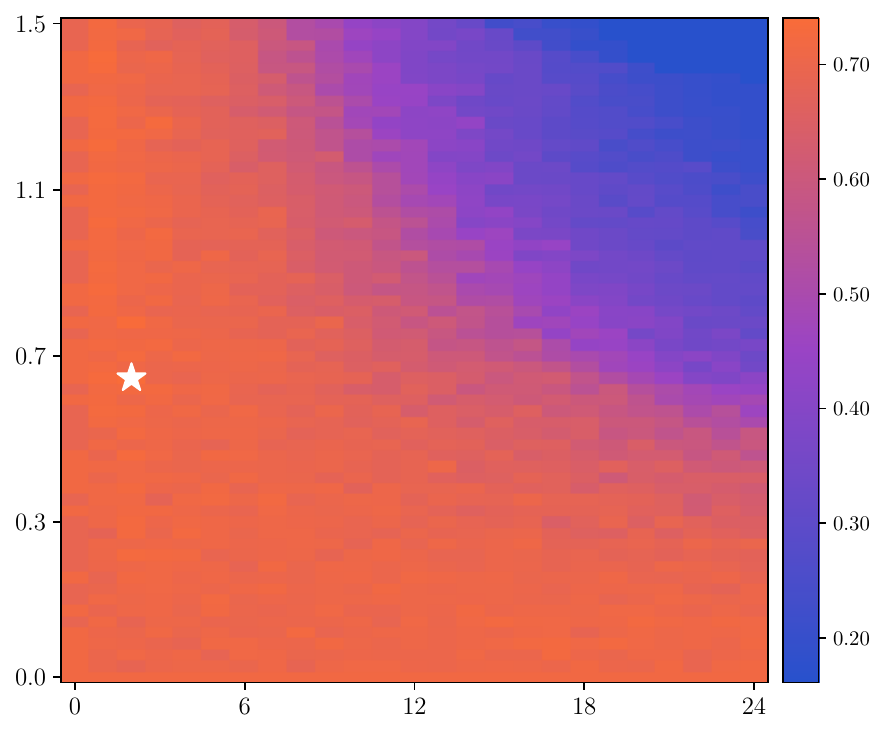}{Olivetti Faces}

  \heatmappair{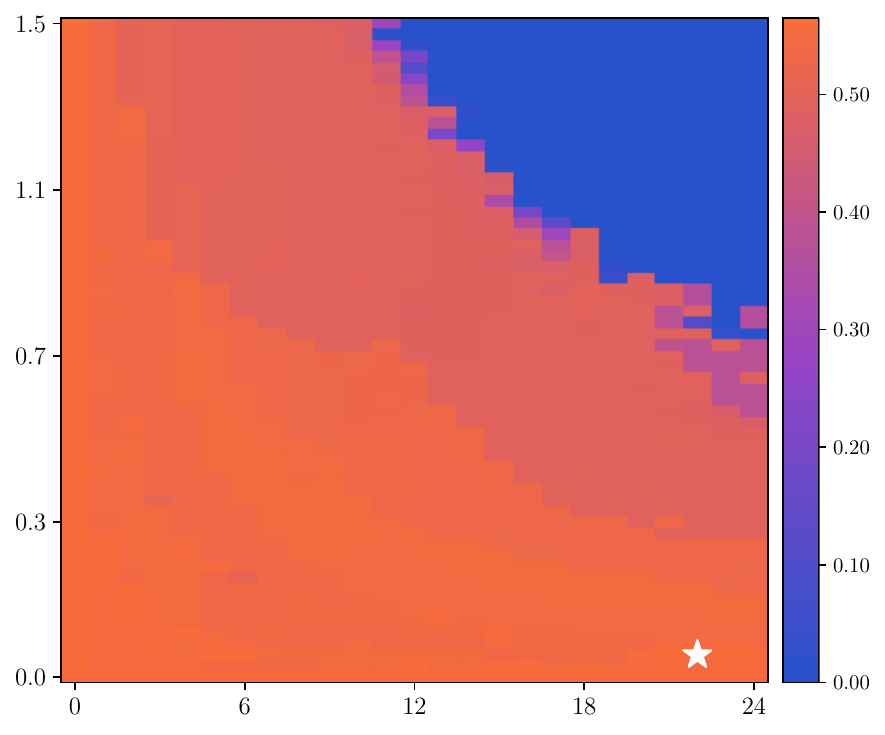}{PH Recognition}
  \hspace{0.66\textwidth} %
  \caption{\textbf{\GSC sensitivity to hyperparameters.} For each dataset, CH and AMI heatmaps are shown over the space of vertex measure parametrization ($t$: x-axis, $\alpha$: y-axis). Shades of orange indicate better clustering, and a white star marks the best configuration for each index, \ie $(t,\alpha)^*_{\textsc{CH}}$ vs $(t,\alpha)^*_{\textsc{AMI}}$. In most cases, the CH landscape aligns well with that of AMI, supporting CH-based unsupervised model selection.}
  \label{fig:heatmaps_sensitivity}
\end{figure}

\inlinetitle{Consistency with external evaluation}{.}%
To improve the interpretability of the results, the clustering solutions identified as best by the previous unsupervised protocol (Tab.~\ref{tab:ch_clustering_scores}) can be evaluated post hoc using the ground-truth labels that come with each datasets. Tab.~\ref{tab:ami_clustering_scores} shows the corresponding AMI scores. In those selected datasets where CH is a suitable proxy for AMI, \GSC variants achieve the best or near-best AMI scores, validating the unsupervised model selection. On smaller, well-separated datasets (\eg Iris, Wine), symmetrized spectral methods remain competitive, showing their adequacy in symmetric, low-noise settings. However, on graphs with asymmetry, imbalance, or noise (\eg Control Chart), GSC's adaptive vertex measure and unnormalized formulation offer clear benefits. Overall, \GSCun achieves the best average AMI ranking ($2$), followed by SC variants ($2.2$ and $3.1$), \GSCn ($3.7$), DI-SIM$_{\text{C}}$ ($3.80$), DI-SIM$_{\text{L}}$ ($4.7$), DI-SIM$_{\text{R}}$ and lastly DSC$+$. This confirms again the advantages of employing the GDE framework in capturing complex graph structures.

In summary, these experiments highlight the flexibility and robustness of the \GSC framework. The tunable parameters $(t,\alpha)$ enable adaptation to diverse graph topologies, from sparse and directed to dense and symmetric. The observed alignment between internal (CH) and external (AMI) metrics across a variety of datasets supports the practical value of CH-based model selection in fully unsupervised settings, provided that CH is a suitable index for the cluster structure of a dataset. By generalizing Dirichlet energy through flexible vertex measures, \GSC provides a principled and scalable approach to graph clustering that performs reliably across a wide range of conditions. Finally, the disagreement between Tab.~\ref{tab:ch_clustering_scores} and Tab.~\ref{tab:ami_clustering_scores} is not negligible, but it rather highlights the difficulty of unsupervised model selection, and more specifically the mismatch between a considered internal evaluation index and the cluster structure present in a dataset.

\begin{table*}[t]
\scriptsize
\centering
\caption{\textbf{Internal CH-based clustering evaluation.} The CH index of each method's solution. The best result per dataset appears bold and shaded; the estimated hyperparameters are in parentheses.}
\vskip -5pt
\label{tab:ch_clustering_scores}
\begin{adjustbox}{width=\linewidth,center}
\begin{tabular}{lccc|rrrrrr|rr}
\Xhline{1pt}
\multicolumn{4}{c|}{\fontsize{6pt}{6pt}\textbf{DATASETS}} & \multicolumn{6}{c|}{\fontsize{6pt}{6pt}\textbf{COMPETITORS}} & \multicolumn{2}{c}{\fontsize{6pt}{6pt}\textbf{\GSC}}\\[0.25em]
\textbf{Name} & $N$ & $d$ & $k$
& \textbf{SC$_{\text{un}}$} & \textbf{SC$_{\text{n}}$} & \textbf{DSC}$+\bestpar{\gamma}{CH}$
& \textbf{DI-SIM$_{\textup{L}}$}$\bestpar{\tau}{CH}$ & \textbf{DI-SIM$_{\text{R}}$}$\bestpar{\tau}{CH}$ & \textbf{DI-SIM$_{\text{C}}$}$\bestpar{\tau}{CH}$
& \textbf{GSC$_{\text{un}}$}$\bestpar{t,\alpha}{CH}$ & \textbf{GSC$_{\text{n}}$}$\bestpar{t,\alpha}{CH}$ \\[0.25em]
\hline
Iris & 150 & 4 & 3 & 555.67 & 555.67 & 22.71 \hyperp{0.05} & 501.63 \hyperp{0.00} & 501.63 \hyperp{0.00} & 501.63 \hyperp{0.00} & 555.67 \hyperp{00, 0.00} & \bestcell{558.06} \hyperp{07, 0.10} \\
Wine & 178 & 13 & 3 & 70.37 & 70.37 & 70.43 \hyperp{0.15} & 69.82 \hyperp{16.00} & 68.54 \hyperp{0.00} & 70.43 \hyperp{16.00} & 70.37 \hyperp{00, 0.00} & \bestcell{70.44} \hyperp{22, 0.10} \\
WDBC & 569 & 30 & 2 & 257.69 & 257.69 & 110.25 \hyperp{0.85} & 257.29 \hyperp{1.00} & 219.76 \hyperp{1.00} & 256.08 \hyperp{1.00} & 257.69 \hyperp{00, 0.00} & \bestcell{258.74} \hyperp{22, 0.60} \\
Control Chart & 600 & 60 & 6 & 341.50 & 290.57 & 166.90 \hyperp{0.85} & 341.50 \hyperp{1.00} & 335.25 \hyperp{1.00} & 341.50 \hyperp{1.00} & 275.10 \hyperp{03, 0.20} & \bestcell{350.42} \hyperp{00, 0.10} \\
Breast Tissue & 106 & 9 & 6 & 50.66 & 47.79 & 27.23 \hyperp{0.05} & 54.39 \hyperp{0.00} & 54.54 \hyperp{0.00} & 54.31 \hyperp{1.00} & \bestcell{54.78} \hyperp{01, 0.40} & 52.04 \hyperp{10, 0.40} \\
Seeds & 210 & 7 & 3 & 247.31 & 247.31 & 192.67 \hyperp{0.80} & 242.24 \hyperp{16.00} & 230.54 \hyperp{0.00} & 231.10 \hyperp{0.00} & 247.31 \hyperp{00, 0.00} & \bestcell{248.04} \hyperp{03, 0.10} \\
Segmentation & 2310 & 19 & 7 & 418.28 & 435.75 & 10.90 \hyperp{0.20} & 451.27 \hyperp{1.00} & 436.24 \hyperp{2.00} & 448.89 \hyperp{1.00} & 418.28 \hyperp{00, 0.00} & \bestcell{507.94} \hyperp{12, 0.60} \\
MNIST64 & 1082 & 64 & 6 & 139.69 & 139.69 & 56.40 \hyperp{0.20} & \bestcell{159.48} \hyperp{6.00} & 143.06 \hyperp{2.00} & 149.34 \hyperp{2.00} & 139.69 \hyperp{00, 0.00} & 159.46 \hyperp{20, 0.90} \\
Olivetti Faces & 400 & 4096 & 5 & 14.41 & 14.33 & 10.68 \hyperp{0.10} & 13.95 \hyperp{5.00} & 11.08 \hyperp{2.00} & 12.62 \hyperp{5.00} & 14.41 \hyperp{00, 0.00} & \bestcell{14.83} \hyperp{09, 0.20} \\
PH Recognition & 653 & 3 & 6 & 436.85 & 436.85 & 8.08 \hyperp{0.05} & 46.45 \hyperp{1.00} & 38.30 \hyperp{1.00} & 41.65 \hyperp{1.00} & 436.85 \hyperp{00, 0.00} & \bestcell{438.91} \hyperp{04, 0.10} \\
\hline
\hline
\multicolumn{4}{r|}{\textbf{Average rank $\rhd$}} & 3.30 & 3.90 & 7.40 & 4.00 & 5.60 & 4.30 & 3.30 & \bestcell{1.50} \\
\multicolumn{4}{r|}{\textbf{Competitiveness index $\rhd$}} & 0.955 & 0.938 & 0.433 & 0.876 & 0.819 & 0.855 & 0.944 & \bestcell{0.995} \\
\Xhline{1pt}
\end{tabular}
\end{adjustbox}
\end{table*}
\begin{table*}[t]
\scriptsize
\centering
\caption{\textbf{External AMI-based clustering evaluation.} The CH index of each method's solution. The best result per dataset appears bold and shaded; the estimated hyperparameters are in parentheses.}
\vskip -5pt
\label{tab:ami_clustering_scores}
\begin{adjustbox}{width=\linewidth,center}
\begin{tabular}{lccc|rrrrrr|rr}
\Xhline{1pt}
\multicolumn{4}{c|}{\fontsize{6pt}{6pt}\textbf{DATASETS}} & \multicolumn{6}{c|}{\fontsize{6pt}{6pt}\textbf{COMPETITORS}} & \multicolumn{2}{c}{\fontsize{6pt}{6pt}\textbf{\GSC}}\\[0.25em]
\textbf{Name} & $N$ & $d$ & $k$
& \textbf{SC$_{\text{un}}$} & \textbf{SC$_{\text{n}}$} & \textbf{DSC}$+\bestpar{\gamma}{AMI}$
& \textbf{DI-SIM$_{\textup{L}}$}$\bestpar{\tau}{AMI}$ & \textbf{DI-SIM$_{\text{R}}$}$\bestpar{\tau}{AMI}$ & \textbf{DI-SIM$_{\text{C}}$}$\bestpar{\tau}{AMI}$
& \textbf{GSC$_{\text{un}}$}$\bestpar{t,\alpha}{AMI}$ & \textbf{GSC$_{\text{n}}$}$\bestpar{t,\alpha}{AMI}$ \\[0.25em]
\hline
Iris & 150 & 4 & 3 & \bestcell{0.803} & \bestcell{0.803} & 0.352 \hyperp{0.05} & 0.753 \hyperp{0.00} & 0.753 \hyperp{0.00} & 0.753 \hyperp{0.00} & \bestcell{0.803} \hyperp{00, 0.00} & 0.775 \hyperp{07, 0.10} \\
Wine & 178 & 13 & 3 & 0.862 & 0.862 & 0.860 \hyperp{0.15} & 0.833 \hyperp{16.00} & \bestcell{0.877} \hyperp{0.00} & 0.860 \hyperp{16.00} & 0.862 \hyperp{00, 0.00} & 0.846 \hyperp{22, 0.10} \\
WDBC & 569 & 30 & 2 & \bestcell{0.677} & \bestcell{0.677} & 0.328 \hyperp{0.85} & 0.662 \hyperp{1.00} & 0.639 \hyperp{1.00} & \bestcell{0.677} \hyperp{1.00} & \bestcell{0.677} \hyperp{00, 0.00} & 0.624 \hyperp{22, 0.60} \\
Control Chart & 600 & 60 & 6 & 0.806 & 0.767 & 0.629 \hyperp{0.85} & 0.806 \hyperp{1.00} & 0.796 \hyperp{1.00} & 0.806 \hyperp{1.00} & \bestcell{0.840} \hyperp{03, 0.20} & 0.824 \hyperp{00, 0.10} \\
Breast Tissue & 106 & 9 & 6 & 0.476 & 0.446 & 0.325 \hyperp{0.05} & 0.474 \hyperp{0.00} & 0.485 \hyperp{0.00} & \bestcell{0.506} \hyperp{1.00} & 0.478 \hyperp{01, 0.40} & 0.484 \hyperp{10, 0.40} \\
Seeds & 210 & 7 & 3 & \bestcell{0.737} & \bestcell{0.737} & 0.622 \hyperp{0.80} & 0.693 \hyperp{16.00} & 0.712 \hyperp{0.00} & 0.696 \hyperp{0.00} & \bestcell{0.737} \hyperp{00, 0.00} & 0.736 \hyperp{03, 0.10} \\
Segmentation & 2310 & 19 & 7 & 0.639 & \bestcell{0.643} & 0.053 \hyperp{0.20} & 0.639 \hyperp{1.00} & 0.614 \hyperp{2.00} & 0.639 \hyperp{1.00} & 0.639 \hyperp{00, 0.00} & 0.606 \hyperp{12, 0.60} \\
MNIST64 & 1082 & 64 & 6 & 0.907 & 0.907 & 0.483 \hyperp{0.20} & 0.961 \hyperp{6.00} & 0.880 \hyperp{2.00} & 0.956 \hyperp{2.00} & 0.907 \hyperp{00, 0.00} & \bestcell{0.972} \hyperp{20, 0.90} \\
Olivetti Faces & 400 & 4096 & 5 & 0.702 & 0.690 & 0.640 \hyperp{0.10} & 0.696 \hyperp{5.00} & 0.635 \hyperp{2.00} & 0.686 \hyperp{5.00} & 0.702 \hyperp{00, 0.00} & \bestcell{0.706} \hyperp{09, 0.20} \\
PH Recognition & 653 & 3 & 6 & 0.560 & 0.560 & 0.160 \hyperp{0.05} & 0.340 \hyperp{1.00} & 0.296 \hyperp{1.00} & 0.335 \hyperp{1.00} & 0.560 \hyperp{00, 0.00} & \bestcell{0.562} \hyperp{04, 0.10} \\
\hline
\hline
\multicolumn{4}{r|}{\textbf{Average rank $\rhd$}} & 2.30 & 3.10 & 7.60 & 4.70 & 5.30 & 3.80 & \bestcell{2.00} & 3.70 \\
\multicolumn{4}{r|}{\textbf{Competitiveness index $\rhd$}} & 0.980 & 0.968 & 0.591 & 0.928 & 0.904 & 0.937 & \bestcell{0.985} & 0.973 \\
\Xhline{1pt}
\end{tabular}
\end{adjustbox}
\end{table*}

\section{Conclusion}
\label{sec:conclusion}

This work introduced the \emph{generalized Dirichlet energy} (GDE) framework, which extends classical notions of graph smoothness to directed graphs through the use of arbitrary vertex measures and random walk dynamics. This leads to a principled construction of parametrized graph Laplacians that can capture both directional and structural properties without relying on teleportation or artificial symmetrization.
Built on this foundation, we developed \emph{generalized spectral clustering} (\GSC), a method that leverages a parametrized vertex measure $\nu_{t,\alpha}$ to adapt to the geometry of directed graphs. The resulting model enables a smooth interpolation between local/global behavior and sparse/dense connectivity, offering a flexible and theoretically grounded approach to clustering in asymmetric and weakly connected settings.

Empirical evaluations on real-world point-cloud datasets show that \GSC consistently performs well—often outperforming classical and teleportation-based baselines, under both internal and external evaluation metrics. These results highlight the value of modeling diffusion via non-ergodic, structure-aware dynamics driven by carefully chosen vertex measures.

The versatility of the GDE framework opens up multiple avenues for future research, including its integration into semi-supervised learning, graph signal processing, and operator-based methods in non-reversible or directed settings. An especially promising direction is the automated or data-driven selection of $(t, \alpha)$, enabling fully adaptive graph inference across tasks and domains.

\section*{Acknowledgments}
Harry Sevi and Argyris Kalogeratos acknowledge support by the Industrial Analytics and Machine Learning (IdAML) Chair hosted at ENS Paris-Saclay, University Paris-Saclay. Gwendal Debaussart-Joniec was funded by a PhD scholarship of the FMJH, Université Paris-Saclay. Matthieu Jonckheere was funded by the International Centre for Mathematics and Computer Science (CImI) in Toulouse.

\bibliographystyle{tmlr}
\bibliography{main}

\appendix

\section{Appendix}

\subsection{Proof of Proposition~\ref{gde_glap}}\label{proof:gdeglap}
\begin{proof}
  We begin by expanding the Dirichlet form:%
  \begin{align*}
    \mathcal{D}_{\nu}(f)
    &= \sum_{i,j \in \gV} \nu(i) p(i,j) [f(i) - f(j)]^2 \\
    &= \sum_{i,j \in \gV} \nu(i) p(i,j) \left( f(i)^2 + f(j)^2 - 2f(i)f(j) \right) \\
    &= \sum_{i \in \gV} \nu(i) f(i)^2
      + \sum_{j \in \gV} \left( \sum_{i \in \gV} \nu(i) p(i,j) \right) f(j)^2
      - 2 \sum_{i,j \in \gV} \nu(i) p(i,j) f(i) f(j) \\
    &= \sum_{i \in \gV} \nu(i) f(i)^2
      + \sum_{j \in \gV} \xi(j) f(j)^2
      - \left( \sum_{i,j \in \gV} \nu(i) p(i,j) f(i) f(j)
      + \sum_{i,j \in \gV} \nu(j) p(j,i) f(i) f(j) \right) \\
    &= \langle f, \rmD_{\nu} f \rangle
      + \langle f, \rmD_{\xi} f \rangle
      - \langle f, (\rmD_{\nu} \rmP + \rmP^{\top} \rmD_{\nu}) f \rangle \\
    &= \langle f, (\rmD_{\nu + \xi} - \rmD_{\nu} \rmP - \rmP^{\top} \rmD_{\nu}) f \rangle \\
    &= \langle f, \rmL_{\nu} f \rangle \\
    &= \langle f, \rmD_{\nu + \xi}^{-1} \rmL_{\nu} f \rangle_{\nu + \xi} \\
    &= \langle f, \rmL_{\textnormal{RW}}(\nu) f \rangle_{\nu + \xi}
  \end{align*}%
  where we used the identity $\langle f, g \rangle_{\nu + \xi} = \langle f, (\rmD_{\nu} + \rmD_{\xi}) g \rangle$, and the definition of the generalized random-walk Laplacian $\rmL_{\textnormal{RW}}(\nu)$ (see Def.~\ref{def:gen_lap}).
\end{proof}

\subsection{Proof of Proposition~\ref{prop:dirichlet_t}.}\label{proof:dirichlett}

\begin{proof}
  As $\rmP$ is ergodic and $\mu$ is assumed to be a probability distribution, we have that $\nu_t = [\rmP^t]^\top \mu \xrightarrow{} \pi$, let $\epsilon_t = \pi - \nu_t$ be the distance of each element of $\nu_t$ to $\pi$.
  \begin{align}
    \D{\nu_t}{f} &= \sum_{i,j\in\gV }\nu_t(i)p(i,j)[f(i)-f(j)]^2 \nonumber \\
                  &= \sum_{i,j\in\gV }(\pi(i) - \epsilon_t(i) )p(i,j)[f(i)-f(j)]^2 \nonumber \\
                  &= \D{\pi}{f} -  \sum_{i,j \in \gV } \epsilon_t(i) p(i,j)[f(i)-f(j)]^2
  \end{align}
  Since $\norm{\epsilon_t}_1 \to 0$, it is clear that the second term vanishes. This result indicates that the GDE of a graph function $f$, associated to the transition matrix $\rmP$ and a parametrized measure $\nu_t$, is the sum of a quadratic form involving the usual unnormalized Laplacian $\rmL_\pi$ (the Dirichlet energy $\D{\pi}{f}$) and a vanishing term. Note that this vanishing term can also be seen as a quadratic form.
\end{proof}

\subsection{Proof of Proposition~\ref{prop:btl_gde}}\label{proof:btlgde}
\begin{proof}
  We begin by expressing the flow from $ S $ to $ \Bar{S} $:
  \begin{align}
    \q(S,\Bar{S})
    &= \sum_{i \in S,\, j \in \Bar{S}} \nu(i) p(i,j) \nonumber \\
    &= \sum_{i,j \in \gV} \nu(i) p(i,j) \chivec{S}(i) \chivec{\Bar{S}}(j) \nonumber \\
    &= \sum_{i,j \in \gV} \nu(i) p(i,j) \chivec{S}(i) (1 - \chivec{S}(j)) \nonumber \\
    &= \sum_{i,j \in \gV} \nu(i) p(i,j) \chivec{S}(i)
      - \sum_{i,j \in \gV} \nu(i) p(i,j) \chivec{S}(i) \chivec{S}(j) \nonumber \\
    \Rightarrow \quad
    \q(S,\Bar{S})
    &= \sum_{i \in \gV} \nu(i) \chivec{S}(i)
      - \sum_{i,j \in \gV} \nu(i) p(i,j) \chivec{S}(i) \chivec{S}(j).
    \label{eq:qssbar}
  \end{align}
  Similarly, the flow from $ \Bar{S} $ to $ S $ is:
  \begin{align}
    \q(\Bar{S},S)
    &= \sum_{i \in \Bar{S},\, j \in S} \nu(i) p(i,j) \nonumber \\
    &= \sum_{i,j \in \gV} \nu(i) p(i,j) \chivec{\Bar{S}}(i) \chivec{S}(j) \nonumber \\
    &= \sum_{i,j \in \gV} \nu(i) p(i,j) (1 - \chivec{S}(i)) \chivec{S}(j) \nonumber \\
    &= \sum_{i,j \in \gV} \nu(i) p(i,j) \chivec{S}(j)
      - \sum_{i,j \in \gV} \nu(i) p(i,j) \chivec{S}(i) \chivec{S}(j) \nonumber \\
    \Rightarrow \quad
    \q(\Bar{S},S)
    &= \sum_{j \in \gV} \left( \sum_{i \in \gV} \nu(i) p(i,j) \right) \chivec{S}(j)
      - \sum_{i,j \in \gV} \nu(i) p(i,j) \chivec{S}(i) \chivec{S}(j).
    \label{eq:qsbarss}
  \end{align}
  Now consider the Dirichlet form evaluated at the indicator vector $ \chivec{S} $:
  \begin{align*}
    \mathcal{D}_{\nu}(\chivec{S})
    &= \sum_{(i,j) \in \mathcal{E}} \nu(i) p(i,j) \left| \chivec{S}(i) - \chivec{S}(j) \right|^2 \\
    &= \sum_{i \in \gV} \nu(i) \chivec{S}(i)
      + \sum_{j \in \gV} \left( \sum_{i \in \gV} \nu(i) p(i,j) \right) \chivec{S}(j)
      - 2 \sum_{i,j \in \gV} \nu(i) p(i,j) \chivec{S}(i) \chivec{S}(j) \\
    \Rightarrow \quad
    \mathcal{D}_{\nu}(\chivec{S})
    &= \q(S,\Bar{S}) + \q(\Bar{S},S),
    \qquad \text{(by Eqs.~\eqref{eq:qssbar} and~\eqref{eq:qsbarss})}.
  \end{align*}
\end{proof}

\subsection{Proof of Corollary~\ref{prop:btl_gde}}\label{proof:btl_gde}
\begin{proof}
  Let $\pi$ be the stationary distribution of the natural random walk on $\mathcal{G}$, i.e. $\pi(j) = \sum_{i \in \gV} \pi(i) \rmP(i,j)$ for all $j$. From the proof of Prop.~\ref{prop:btl_gde}, we have:
  \begin{align*}
    \q(S, \Bar{S}) - \q(\Bar{S}, S)
    &= \sum_{i \in \gV} \pi(i) \chivec{S}(i) - \sum_{j \in \gV} \left( \sum_{i \in \gV} \pi(i) \rmP(i,j) \right) \chivec{S}(j) \\
    &= \sum_{i \in \gV} \pi(i) \chivec{S}(i) - \sum_{j \in \gV} \pi(j) \chivec{S}(j) = 0.
  \end{align*}
  Thus, $\q(S, \Bar{S}) = \q(\Bar{S}, S)$, and by Prop.~\ref{prop:btl_gde}, we conclude that:
  \[
    \D{\pi}{\chivec{S}} = \q(S, \Bar{S}) + \q(\Bar{S}, S) = 2\q(S, \Bar{S}).
  \]
\end{proof}

\end{document}